\documentclass[12pt]{article}
\usepackage{amsmath,amssymb}
\usepackage{bm,color,comment}
\usepackage{graphicx}
\usepackage{natbib}
\usepackage{url} 

\newcommand{\blind}{0}

\newtheorem{Thm}{Theorem}[section]
\newtheorem{Rem}[Thm]{Remark}

\newtheorem{Asm}[Thm]{Assumption}

\newtheorem{Coro}[Thm]{Corollary}
\newtheorem{claim}[Thm]{Claim}

\begin{document}

\def\spacingset#1{\renewcommand{\baselinestretch}%
{#1}\small\normalsize} \spacingset{1}


\if0\blind
{
  \title{\bf Privacy-Accuracy Trade-offs in High-Dimensional LASSO under Perturbation Mechanisms}
  \author{Ayaka Sakata\thanks{
    The authors gratefully acknowledge JSPS KAKENHI
(22H05117) and JST PRESTO (JPMJPR23J4).}\hspace{.2cm}\\
    Department of Information Science, Ochanomizu University\\
    RIKEN center for AIP\\
    \and 
    Haruka Tanzawa \\
    Department of Information Science, Ochanomizu University}
  \maketitle
} \fi

\if1\blind
{
  \bigskip
  \bigskip
  \bigskip
  \begin{center}
    {\LARGE\bf Title}
\end{center}
  \medskip
} \fi

\bigskip
\begin{abstract}
We study privacy-preserving sparse linear regression in the high-dimensional regime, focusing on the LASSO estimator. We analyze two widely used mechanisms for differential privacy: output perturbation, which injects noise into the estimator, and objective perturbation, which adds a random linear term to the loss function.

Using approximate message passing (AMP), we characterize the typical behavior of these estimators under random design and privacy noise. To quantify privacy, we adopt typical-case measures, including the on-average KL divergence, which admits a hypothesis-testing interpretation in terms of distinguishability between neighboring datasets.

Our analysis reveals that sparsity plays a central role in shaping the privacy-accuracy trade-off: stronger regularization can improve privacy by stabilizing the estimator against single-point data changes. We further show that the two mechanisms exhibit qualitatively different behaviors. In particular, for objective perturbation, increasing the noise level can have non-monotonic effects, and excessive noise may destabilize the estimator, leading to increased sensitivity to data perturbations.

Our results demonstrate that AMP provides a powerful framework for analyzing privacy-accuracy trade-offs in high-dimensional sparse models.
\end{abstract}

\noindent%
\vfill

\newpage
\spacingset{1.75} 

\section{Introduction}
\label{sec:Introduction}

Differential privacy (DP) has become a foundational concept in modern data analysis, providing formal guarantees that protect
whether an individual’s data is included in a dataset \citep{dwork2006calibrating}.
Its practical relevance is evidenced by its deployment in production systems by large-scale tech companies and governmental institutions. 
For example, the U.S. Census Bureau employed differential privacy for the 2020 decennial census data release, marking one of the first large-scale governmental uses of DP \citep{kenny2021use}. 
Similarly, technology platforms use DP to protect individual contributions in aggregate statistics derived from user behavior, recommendation systems \citep{mullner2023differential}, and natural language processing pipelines \citep{klymenko-etal-2022-differential}.
In addition, there is increasing demand for privacy-preserving analysis in domains such as healthcare, genomics, and finance, where both the dimensionality of data and the risk of privacy leakage are high.

A common approach to achieve differential privacy is to randomize the process of parameter estimation via a \textit{randomized mechanism}.
Several such mechanisms have been proposed, including output perturbation, where noise is added directly to the estimated parameters \citep{dwork2006calibrating}, and objective perturbation, where noise is injected into the loss function before optimization \citep{chaudhuri2011differentially}.
A fundamental challenge in all such approaches is balancing the privacy-utility trade-off \citep{dwork2014algorithmic}: stronger privacy guarantees typically require more randomness, which can deteriorate statistical accuracy and model interpretability.

In sparse estimation problems such as LASSO, output perturbation can destroy sparsity patterns and deteriorate statistical efficiency \citep{chaudhuri2008privacy}.
A natural approach is to first estimate the support of the parameters and then apply perturbation only to the selected components. However, such two-stage procedures require privatizing the support selection step itself, which consumes additional privacy budget by the composition property of differential privacy \citep{dwork2014algorithmic}.
In contrast, objective perturbation implicitly accounts for the regularization structure of the estimator, and can better preserve sparsity without an explicit support selection step.
Despite this appeal, existing theoretical analyses of perturbation-based mechanisms face fundamental limitations in high-dimensional sparse settings; see Section~\ref{sec:related} for a detailed discussion.

To better understand these interactions, we focus on the high-dimensional regime where both the number of samples and parameters grow to infinity with a fixed ratio $n/p \to \alpha \ll\infty$. In such settings, worst-case DP guarantees can be overly conservative, as they are dominated by rare pathological configurations that do not reflect typical data realizations. We instead adopt a typical-case perspective, analyzing privacy leakage under random design and noise via the on-average KL divergence (OnAveKL) \citep{wang2016average}, which admits an operational interpretation in terms of hypothesis testing: a small OnAveKL implies that even an optimal adversary cannot reliably infer the presence of an individual data point.

Building on this formulation, we analyze the high-dimensional LASSO under both output perturbation and objective perturbation using approximate message passing (AMP). AMP provides an asymptotic characterization of high-dimensional estimators under random design, enabling a typical-case analysis of generalization error and privacy leakage jointly, without relying on restricted eigenvalue assumptions or worst-case sensitivity bounds.

\subsection{Related Works}
\label{sec:related}

We review related work from the perspectives of private estimation, empirical risk minimization, and high-dimensional asymptotic analysis, with a focus on the gap between worst-case guarantees and typical-case behavior in high-dimensional regimes.

\paragraph{Perturbation-based mechanisms for private estimation.}
Several works have developed perturbation-based mechanisms for differentially private statistical estimation \citep{chaudhuri2011differentially, smith2011privacy}. However, existing analyses rely on assumptions incompatible with sparse estimation: strong convexity excludes $\ell_1$-type penalties \citep{chaudhuri2011differentially}, and asymptotic normality fails for sparse estimators \citep{smith2011privacy}. \citet{kifer2012private} extended objective perturbation to non-differentiable regularizers via a two-stage approach, though their analysis requires restricted strong convexity. \citet{talwar2015nearly} achieved nearly optimal excess risk of $\tilde{O}(1/n^{2/3})$ for private LASSO without such structural assumptions, but these results are established in a finite-sample regime and do not characterize behavior in the proportional high-dimensional limit $p/n \to \delta > 0$, nor the interplay between privacy noise, regularization, and generalization performance.

\paragraph{Private empirical risk minimization.}
In the broader context of private empirical risk minimization, \citet{bassily2014private} established minimax optimal rates for convex ERM under only Lipschitz and boundedness assumptions, providing matching upper and lower bounds for both $(\epsilon, 0)$- and $(\epsilon, \delta)$-differential privacy. 
While these guarantees are foundational, they are worst-case in nature and do not characterize behavior in the proportional high-dimensional limit.
Related developments include methods that exploit additional structure to mitigate the dimensional dependence inherent in private optimization. 
For example, \citet{wang2018empirical} shows that, under smoothness or model assumptions, the sample complexity of ERM under local differential privacy can be significantly improved via polynomial approximation. 
Similarly, \citet{zhou2020bypassing} proposes a differentially private SGD algorithm that leverages low-dimensional gradient structure to reduce the dependence on the ambient dimension. 
While these approaches rely on additional structural assumptions to improve upon worst-case bounds, their analyses are still primarily framed in worst-case terms and do not capture the typical-case behavior in high-dimensional regimes.

\paragraph{On-average privacy measures.}
On-average privacy measures such as OnAveKL \citep{wang2016average} have been proposed to capture typical-case privacy leakage, admitting an operational interpretation in terms of hypothesis testing and membership inference. However, existing studies have largely focused on low-dimensional or unregularized settings, leaving open the question of how high dimensionality, sparsity, and regularization jointly affect on-average privacy, particularly in regimes where precise asymptotic characterization becomes possible.

\paragraph{High-dimensional analysis via AMP.}
Approximate message passing (AMP) provides a powerful framework for the precise asymptotic characterization of high-dimensional estimators under random design through state evolution \citep{donoho2009message, bayati2011dynamics}. 
It has been successfully applied to analyze the typical-case performance of regularized estimators such as the LASSO in the proportional high-dimensional regime. 
However, its application to privacy-constrained estimation problems remains limited. 
In this work, we bridge this gap by combining AMP with OnAveKL, enabling a typical-case analysis that jointly captures the effects of privacy mechanisms and statistical estimation in high-dimensional sparse regression, and quantitatively characterizes the trade-off between privacy and generalization.

\subsection{Our contribution}

In this work, we analyze differentially private sparse linear regression in the high-dimensional regime, focusing on the LASSO estimator under both output perturbation and objective perturbation. Our main contributions are summarized as follows:

\begin{itemize}

\item \textbf{Asymptotic characterization of perturbation mechanisms:}
We derive an explicit asymptotic characterization of the LASSO estimator under both output perturbation and objective perturbation in the high-dimensional limit. Using approximate message passing, we reduce the high-dimensional problem to a scalar equivalent model, which captures the combined effects of sparsity, noise injection, and sample complexity.

\item \textbf{Closed-form characterization of typical-case accuracy and privacy:}
We obtain explicit formulas for both prediction accuracy and the on-average KL divergence (OnAveKL) in the high-dimensional limit. Our results characterize the dependence of both quantities on the noise variance and the sparsity level induced by regularization.

\item \textbf{Comparison between output and objective perturbation:}
We establish a quantitative comparison between output perturbation and objective perturbation within a unified framework. Our analysis reveals that objective perturbation can achieve smaller OnAveKL in regimes when the support of the estimator remains stable under the replacement of a single data point.

\item \textbf{Non-monotonic effects of objective noise on privacy and stability:}
Our analytical characterization provides practical guidance for selecting the noise level to balance prediction accuracy and privacy. In particular, we identify regimes where increasing noise leads to non-monotonic effects on stability and privacy under objective perturbation.

\end{itemize}

Taken together, our results provide a quantitative characterization of privacy–accuracy trade-offs in high-dimensional sparse estimation, complementing existing worst-case analyses.

\section{Problem setting}

We consider sparse linear regression under differential privacy and study two widely used randomization mechanisms: output perturbation and objective perturbation. In the following, we introduce these two mechanisms.

\subsection{Private LASSO via Objective Perturbation}

We consider a linear regression model to estimate a sparse parameter vector 
$\bm{\beta}^{(0)} \in \mathbb{R}^p$ from a dataset 
${\cal D} = \{(\bm{x}_\mu, y_\mu)\}_{\mu=1}^n$, 
where
\begin{align}
y_\mu = \bm{x}_\mu^\top \bm{\beta}^{(0)} + \xi_\mu,
\label{eq:generative_process}
\end{align}
and the noise variables $\xi_\mu$ are i.i.d. Gaussian with 
$\xi_\mu \sim \mathcal{N}(0, \sigma^2)$.

Under the objective perturbation mechanism, noise is injected directly into the function to be optimaized. The LASSO estimator is defined as
\begin{align}
\widehat{\bm{\beta}}({\cal D}, \bm{\eta})
= \mathop{\mathrm{argmin}}_{\bm{\beta}} 
\left\{
\frac{1}{2} \|\bm{y} - X \bm{\beta}\|_2^2
+ \lambda \|\bm{\beta}\|_1
+ \bm{\eta}^\top \bm{\beta}
\right\},
\label{eq:lasso_obj_perturb}
\end{align}
where $X = (\bm{x}_1, \ldots, \bm{x}_n)^\top \in \mathbb{R}^{n \times p}$ and $\bm{\eta}\in\mathbb{R}^p$ is a random noise vector drawn from a distribution $P_\eta(\bm{\eta})$.
The magnitude of $\bm{\eta}$ controls the strength of privacy protection and affects both estimation accuracy and information leakage.

\subsection{Private LASSO via Output Perturbation}

Under the output perturbation mechanism, noise is added directly to the estimator. The resulting estimator is defined as
\begin{align}
    \widehat{\bm{\beta}}({\cal D},\bm{\eta})
    =
    \widehat{\bm{\beta}}({\cal D},\bm{0})+\bm{\eta},
\end{align}
where $\widehat{\bm{\beta}}({\cal D},\bm{0})$ denotes the standard LASSO estimator obtained in the absence of privacy noise, defined by \eqref{eq:lasso_obj_perturb} with $\bm{\eta}=\bm{0}$.


\subsection{Evaluation Criteria: Accuracy and Privacy}

To investigate the trade-off between utility and privacy, we evaluate the estimator using two complementary criteria.

\paragraph{(1) Prediction Accuracy.}
The prediction performance is evaluated using the expected generalization error:
\begin{align}
E_{\mathrm{gen}}(\bm{\eta})
= 
\mathbb{E}_{{\cal D},\,\bm{x}_{\mathrm{new}},\,\xi_{\mathrm{new}}}
\left[
\left(
y_{\mathrm{new}} 
- 
\bm{x}_{\mathrm{new}}^\top 
\widehat{\bm{\beta}}({\cal D}, \bm{\eta})
\right)^2
\right],
\end{align}
where $\bm{x}_{\mathrm{new}}$ is a test sample not used in training, and 
$y_{\mathrm{new}} = \bm{x}_{\mathrm{new}}^\top \bm{\beta}^{(0)} + \xi_{\mathrm{new}}$ with observation noise
$\xi_{\mathrm{new}}$.

The training error is similarly defined as
\begin{align}
E_{\mathrm{train}}(\bm{\eta})
=
\frac{1}{n}\,
\mathbb{E}_{{\cal D}}
\left[
\| 
\bm{y} 
- 
X \widehat{\bm{\beta}}({\cal D}, \bm{\eta})
\|_2^2
\right].
\end{align}
Comparing the generalization and training errors provides insight into how much the estimator relies on the training data.
In particular, the increase of the generalization error relative to the training error can serve as an indicator of potential privacy leakage, since stronger dependence on individual samples may reveal information about the training dataset \citep{yeom2018privacy, wu2019generalization}.

For output perturbation, when $\bm{\eta}$ is independent of the dataset and the test sample, the generalization and training errors become
\begin{align}
E_{\mathrm{gen}}(\bm{\eta})
&=
E_{\mathrm{gen}}(\bm{0})
+
\mathbb{E}
\left[
(\bm{x}_{\mathrm{new}}^\top \bm{\eta})^2
\right],\label{eq:gen_output}\\
E_{\mathrm{train}}(\bm{\eta})
&=
E_{\mathrm{train}}(\bm{0})
+
\frac{1}{n}
\mathbb{E}_{{\cal D}}
\left[
\|
X\bm{\eta}
\|_2^2
\right].
\end{align}
Hence, output perturbation always increases both the generalization and training errors compared with the noiseless estimator.

In contrast, for objective perturbation the effect of privacy noise on the estimator is nontrivial, since the noise enters through the optimization problem itself. 
Analyzing its impact on these errors is therefore one of the main objectives of this work.

\paragraph{(2) Privacy Measure: On-Average KL-Privacy.}

To quantify privacy, we adopt an information-theoretic perspective based on the distinguishability between neighboring datasets. Consider an adversary who observes the estimator and attempts to determine whether it was computed from a dataset ${\cal D}$ or from a neighboring dataset ${\cal D}^\prime$ that differs in a single sample. This naturally leads to a binary hypothesis testing problem.

Due to the injected privacy noise, the estimator in \eqref{eq:lasso_obj_perturb} is a random variable even for a fixed dataset ${\cal D}$. We thus define the distribution of the estimated parameters as
\begin{align}
\mathbb{P}(\bm{\beta}\,|\,{\cal D})
= \int d\bm{\eta}\, P_\eta(\bm{\eta})\,
\delta\!\left(\widehat{\bm{\beta}}({\cal D}, \bm{\eta}) - \bm{\beta}\right).
\end{align}
Under the two hypotheses ${\cal D}$ and ${\cal D}^\prime$, the adversary observes samples from $\mathbb{P}(\bm{\beta}|{\cal D})$ or $\mathbb{P}(\bm{\beta}|{\cal D}^\prime)$.
In this setting, classical results in hypothesis testing imply that the optimal error exponent for distinguishing the two distributions is governed by the Kullback-Leibler (KL) divergence between them. In particular, by Stein's lemma, a smaller KL divergence implies that even the most powerful test cannot reliably distinguish whether a given data point was included in the dataset. This provides a direct operational interpretation of KL-based privacy measures.

Motivated by this perspective, we define the On-Average KL-Privacy as
\begin{align}
\mathrm{OnAvgKL}
= \frac{1}{n}\sum_{\mu=1}^n\mathbb{E}_{{\cal D}, {\cal D}_\mu'}
\left[
\mathrm{KL}\!\left(
\mathbb{P}(\bm{\beta}|{\cal D})
\big\|
\mathbb{P}(\bm{\beta}|{\cal D}_\mu')
\right)
\right],
\label{eq:def_OnAveKL}
\end{align}
where ${\cal D}_\mu^\prime$ denotes a dataset differing from ${\cal D}$ only in its $\mu$-th sample.
The quantity measures how much the output distribution of the estimator changes, on average, when a single sample is replaced. A small value of OnAvgKL indicates that the two distributions are nearly indistinguishable, implying that the inclusion or exclusion of an individual sample has little influence on the estimator.
This KL-based measure applies to both output perturbation and objective perturbation, allowing a unified comparison of their privacy properties through the same information-theoretic criterion.

\section{Approximate Message Passing algorithm}

\subsection{Bayesian formulation and MAP estimator}
\label{sec:MAP}

We reformulate the LASSO estimator within a Bayesian framework, which provides a unified starting point for analyzing both objective perturbation and output perturbation mechanisms. 
We define the posterior distribution as
\begin{align}
P_\tau(\bm{\beta} \mid \mathcal{D}; \bm{\eta}, \lambda)
= \frac{1}{Z_\tau(\mathcal{D}; \bm{\eta}, \lambda)}
\exp \Bigg\{
-\tau \Big(
\frac{1}{2} \|\bm{y}-X\bm{\beta}\|_2^2
+ \lambda \|\bm{\beta}\|_1
+ \bm{\eta}^\top \bm{\beta}
\Big)
\Bigg\},
\label{eq:def_posterior}
\end{align}
where $Z_\tau(\mathcal{D};\bm{\eta},\lambda)$ denotes the normalization constant and $\lambda>0$ is the regularization parameter, and $\tau$ controls the sharpness of the posterior.
In the limit $\tau \to \infty$, the posterior concentrates around its mode, and the estimator converges to the corresponding maximum a posteriori (MAP) solution, given by eq. \eqref{eq:lasso_obj_perturb}.
This posterior formulation provides a starting point for the microscopic AMP-based approximation presented in the following sections.


\subsection{Assumptions on data generative process and privacy noise}

We assume that data are generated according to \eqref{eq:generative_process} under the following assumptions, which allow for a tractable analysis in the high-dimensional limit:
\paragraph{Setting D (Data model)}
\begin{enumerate}
\item \textbf{Predictor matrix.}
      The entries of $X$ are i.i.d. Gaussian:
$X_{\mu i} \sim \mathcal{N}(0,1/p)$, so that $\mathbb{E}\left[\sum_{i=1}^p X_{\mu i}^2\right] = 1$.

\item \textbf{Ground truth coefficients.}
      The true signal follows a Bernoulli-Gaussian prior:
      \[
      \beta^{(0)}_i \sim (1-\rho)\delta_0 + \rho\,\mathcal{N}(0,\sigma_\beta^2),
      \]
      where $\rho\in[0,1]$ denotes the fraction of nonzero components, and $\delta_0$ is the Dirac delta at zero.

\item \textbf{Observation noise.}
      Measurement noise is i.i.d. Gaussian:
      $\xi_\mu \sim \mathcal{N}(0,\sigma_\xi^2)$.
\end{enumerate}

In addition to the data-generating assumptions above, we specify the distribution of the privacy noise as follows:

\paragraph{Setting P (Privacy noise model)}
\begin{enumerate}
\item \textbf{Independence across coordinates:}
      The entries $\{\eta_i\}_{i=1}^p$ are mutually independent.

\item \textbf{Independence from the data:}
      $\bm{\eta}$ is independent of $(X,\bm{y})$.

\item \textbf{Identical marginal distribution:}
      All coordinates $\eta_i$ share the same marginal distribution.

\item \textbf{Finite variance:}
      $\mathbb{E}[\eta^2] < \infty$.
\end{enumerate}

Under Setting~\textbf{P}, for any pseudo-Lipschitz function $\phi$,
\begin{align}
\frac{1}{p}\sum_{i=1}^p \phi(\eta_i)
\xrightarrow{p}
\mathbb{E}_{\eta}[\phi(\eta)],
\label{eq:LLN}
\end{align}
where $\xrightarrow{p}$ denotes convergence in probability as $p\to\infty$.

Thus, in the high-dimensional limit, the effect of the perturbation can be captured through deterministic scalar summary statistics (e.g., its variance) when evaluating pseudo-Lipschitz observables, rather than its full realization.

\subsection{Approximate Message Passing Algorithm}

To analyze the high-dimensional behavior of the estimator, we employ approximate message passing (AMP), which provides a tractable approximation to the marginal posterior distributions in large random systems. This approach enables a  characterization of both estimation performance and distributional properties relevant for privacy analysis.

The message passing framework aims to approximate the marginal posterior of each component:
\begin{align}
p_{\tau,i}(\beta_i|{\cal D},\bm{\eta}):=\int d\bm{\beta}_{\setminus i}P_{\tau}(\bm{\beta}|{\cal D},\bm{\eta}),
\end{align}
and corresponding posterior mean at $\tau\to\infty$
\begin{align}
    \widehat{\beta}_i({\cal D},\bm{\eta})=\lim_{\tau\to \infty}\int d\beta_i~\beta_i p_{\tau,i}(\beta_i|{\cal D},\bm{\eta}).
\end{align}
The marginal distribution is approximated via an iterative message passing procedure. At iteration $t$, the approximate marginal $p_{\tau,i}^{(t)}$ is constructed as
\begin{align}
p_{\tau,i}^{(t)}(\beta_i)
\propto
\exp\left(-\tau\lambda|\beta_i| - \tau\eta_i\beta_i\right)
\prod_{\mu \in \bm{n}} \widetilde{\pi}_{\mu \to i}^{(t)}(\beta_i),
\label{eq:marginal_bp}
\end{align}
where, for notational simplicity, we omit the dependence on $({\cal D}, \bm{\eta})$.
The messages $\widetilde{\pi}_{\mu \to i}^{(t)}(\beta_i)$, defined for $\mu \in \{1,\ldots,n\}$ and $i \in \{1,\ldots,p\}$, are updated recursively according to
\begin{align}
    \pi_{i\to\mu}^{(t+1)}(\beta_i)&\propto\exp\left(-\tau\lambda|\beta_i|-\tau\eta_i\beta_i\right)\prod_{\nu\in\bm{n}\setminus \mu}\widetilde{\pi}_{\nu\to i}^{(t)}(\beta_i)\label{eq:def_output_message}\\
    \widetilde{\pi}_{\mu\to i}^{(t)}(\beta_i)&\propto\int d\bm{\beta}_{\setminus i}\exp\left\{-\frac{\tau}{2}(y_\mu-\bm{x}^\top_\mu\bm{\beta})^2\right\}\prod_{j\in\bm{p}\setminus i}\pi_{j\to\mu}^{(t)}(\beta_j),\label{eq:def_input_message}
\end{align}
where $\bm{n} = \{1,\ldots,n\}$ and $\bm{p} = \{1,\ldots,p\}$.
The expression \eqref{eq:marginal_bp} becomes exact after sufficient iterations when the underlying factor graph is a tree. For loopy graphs, the same update rules can still be applied, in which case the procedure can be interpreted as a form of variational inference.

\subsubsection{Gaussian Approximation}

The update equations \eqref{eq:def_output_message} and 
\eqref{eq:def_input_message} involve $2np$ messages, 
which are computationally intractable in high dimensions. 
To obtain a tractable algorithm in the large-system limit 
($n,p \to \infty$ with $\alpha = n/p$ fixed), 
we introduce the following standard assumptions.

\begin{Asm}
The entries of the predictor matrix $X$ are $O(p^{-1/2})$ so that 
$\|\bm{x}_\mu\|_2^2 = O(1)$.
\label{assum:scaling}
\end{Asm}
\begin{Asm}
The empirical correlations between different predictors vanish 
as $p \to \infty$, i.e., the off-diagonal elements of $X^\top X$ 
are $O(p^{-1/2})$.
\label{assum:weak_corr}
\end{Asm}
\begin{Asm}
The factor graph associated with $X$ is dense, 
with $O(p)$ edges connected to each factor node.
\label{assum:dense}
\end{Asm}
Predictor matrices generated according to Setting~D satisfy Assumptions~\ref{assum:scaling}--\ref{assum:dense}.
Under Assumptions \ref{assum:scaling}--\ref{assum:dense}, 
the messages $\{\widetilde{\pi}_{\mu\to i}\}$ 
become asymptotically Gaussian due to the central limit effect 
\cite{krzakala2012probabilistic,sakata2023prediction}. 
Therefore, it suffices to track only their means and variances:
\begin{align}
\widehat{\beta}_{i\to\mu}^{(t)}&=\int d\beta_i \beta_i\pi_{i\to\mu}^{(t)}(\beta_i)\\
s_{i\to\mu}^{(t)}&=\tau\int d\beta_i (\beta_i-\widehat{\beta}_{i\to\mu}^{(t)})^2\pi_{i\to\mu}^{(t)}(\beta_i)
\label{eq:s_def}\\
\widetilde{\beta}_{\mu\to i}^{(t)}&=\int d\beta_i \beta_i\widehat{\pi}_{\mu\to i}^{(t)}(\beta_i)\\
\widetilde{s}_{\mu\to i}^{(t)}&=\tau\int d\beta_i (\beta_i-\widetilde{\beta}_{\mu\to i}^{(t)})^2\widetilde{\pi}_{\mu\to i}^{(t)}(\beta_i),\label{eq:s_tilde_def}
\end{align}
The resulting algorithm that updates these means and variances 
for arbitrary likelihood and prior distributions 
is known as approximate message passing (AMP).

Within AMP, the marginal posterior of each coefficient at iteration $t$ reduces to a scalar distribution. Specifically, the scalar marginal distribution for $i$-th component is given by
\begin{align}
{\cal P}_\tau(\beta;\Sigma^{(t)},\mathrm{m}_i^{(t)}-\eta\Sigma^{(t)})
&=
\frac{1}{{\cal Z}_\tau(\Sigma^{(t)},\mathrm{m}^{(t)}_i-\eta_i\Sigma^{(t)})}
\exp\left\{
-\frac{\tau\left(\beta-\left(\mathrm{m}_i^{(t)}-\eta_i\Sigma^{(t)}\right)\right)^2}{2\Sigma^{(t)}}
-\tau\lambda|\beta|
\right\},
\label{eq:AMP_one-body_dist}
\end{align}
where ${\cal Z}_\tau(\Sigma,\mathrm{m}-\eta\Sigma)$ is the normalization constant. The mean parameter $\mathrm{m}_i$ and 
variance parameter $\Sigma$ are given by
\begin{align}
\mathrm{m}_{i}^{(t)}&={\Sigma}^{(t)}\sum_{\mu=1}^n\left\{
X_{\mu i}\frac{y_\mu-\widehat{y}_\mu^{\setminus\mu(t)}}{1+{s_\theta^{(t)}}}
+\frac{X_{\mu i}^{2}\widehat{\beta}_{i}^{(t)}}{1+{s_\theta^{(t)}}}\right\}\\
{\Sigma}^{(t)}&=\left(\frac{\alpha}{1+{s_\theta^{(t)}}}\right)^{-1},
\end{align}
where 
\begin{align}
    \widehat{y}_\mu^{\setminus\mu}&=\sum_{i=1}^pX_{\mu i}\widehat{\beta}_{i\to\mu}\\
    s_\theta^{(t)}&=\frac{1}{p}\sum_{i=1}^ps_i^{(t)}
\end{align}
and we set
\begin{align}
    s_i^{(t)}=\tau \int d\beta_i(\beta_i-\widehat{\beta}_i^{(t)})^2{\cal P}_\tau(\beta;\Sigma^{(t)},\mathrm{m}_i^{(t)}-\eta_i\Sigma^{(t)}).
\end{align}
Using the marginal distribution, we obtain the estimate
\begin{align}
    \widehat{\beta}_i^{(t)}&=\int d\beta_i \beta_i{\cal P}_\tau(\beta;\Sigma^{(t)},\mathrm{m}_i^{(t)}-\eta_i\Sigma^{(t)}).
\end{align}
We note that the dependence on the dataset appears in $\{\mathrm{m}_i\}$ and $\Sigma$. In addition, for the $i$-th marginal, the privacy noise $\eta_i$ appears explicitly, while the remaining noise components $\{\eta_j\}_{j \neq i}$ enter implicitly through the parameters $\mathrm{m}_i$ and $\Sigma$.

\noindent{\bf $\tau\to\infty$ limit}

In the limit $\tau\to\infty$, we apply Laplace's approximation 
with respect to $\beta$ to the distribution 
\eqref{eq:AMP_one-body_dist}. 
As a result, expectations with respect to this distribution reduce to
\begin{align}
\widehat{\beta}_i^{(t)}&=\mathbb{M}(\Sigma^{(t)},\mathrm{m}_i^{(t)}-\eta_i\Sigma^{(t)})\\
s_i^{(t)}&=\mathbb{V}(\Sigma^{(t)},\mathrm{m}_i^{(t)}-\eta_i\Sigma^{(t)}),
\end{align}
where 
\begin{align}
\nonumber
    \mathbb{M}(\Sigma,\mathrm{m}-\eta\Sigma)&=\mathop{\mathrm{argmin}}_\beta\left\{\frac{(\beta-(\mathrm{m}-\eta\Sigma))^2}{2\Sigma}-\lambda|\beta|\right\}\\
    &=\left(\mathrm{m}-\eta\Sigma-\mathrm{sgn}(\mathrm{m}-\eta\Sigma\right)\lambda\Sigma)\mathbb{I}(|\mathrm{m}-\eta\Sigma|>\lambda\Sigma)\label{eq:AMP_M}\\
    \mathbb{V}(\Sigma,\mathrm{m}-\eta\Sigma)&={\Sigma}\mathbb{I}(|\mathrm{m}-\eta\Sigma|>\lambda\Sigma),
\end{align}
and $\mathbb{I}(a)$ denotes the indicator function, 
which returns 1 if $a$ is true and 0 otherwise.

Equation~\eqref{eq:AMP_M} corresponds to a soft-thresholding operator. The privacy noise shifts the mean parameter to $\mathrm{m}-\eta\Sigma$, thereby affecting whether the coefficient is set to zero or remains active.

\noindent{\bf Estimate under LOO sample}

Intuitively, the output message $\pi_{i\to \mu}(\beta_i)$ 
represents the marginal distribution of $\beta_i$ 
before incorporating the observation $y_\mu$ associated with $\bm{x}_\mu$, 
that is, under the leave-one-out (LOO) dataset 
${\cal D}_{\setminus\mu}$ in a tree-structured factor graph, where
${\cal D}_{\setminus\mu}
=\left\{(\bm{x}_{\nu},y_\nu)\mid \nu=1,\dots,n,\ \nu\neq\mu \right\}$ \citep{koller2009probabilistic}.
Although regression problems with dense predictor matrices 
do not form tree graphs, 
we adopt the following approximation:
\begin{Asm}
The mean of the output message $\widehat{\beta}_{i\to\mu}$ 
coincides with the estimator of the $i$-th parameter 
computed from the leave-one-out dataset 
${\cal D}_{\setminus\mu}$.
\label{asm:LOO}
\end{Asm}

Under the Assumption \ref{asm:LOO}, 
the estimate and its rescaled variance under the LOO dataset are given by
\begin{align}
\widehat{\beta}^{(t+1)}&=\mathbb{M}
\left({\Sigma}^{(t)},\mathrm{m}^{(t)}_{i\to\nu}-\eta\Sigma^{(t)}\right),\\
{s}^{(t+1)}_{i\to\nu}&=\mathbb{V}
\left({\Sigma}^{(t)},\mathrm{m}^{(t)}_{i\to\nu}-\eta\Sigma^{(t)}\right).\label{eq:output_stat}
\end{align}
From a perturbative expansion, we obtain the relation \citep{krzakala2012probabilistic,sakata2023prediction},
\begin{align}
\mathrm{m}^{(t)}_{i\to\mu}
&=\mathrm{m}_i^{(t)}-X_{\mu i}\frac{y_\mu-\widehat{y}_\mu^{\setminus\mu(t)}}{1+s_\theta^{(t)}}\Sigma^{(t)}
\label{eq:m_cavity_vs_m}
\end{align}

\section{Effect of Privacy Noise on errors and sparsity}

For simplicity, we set $\sigma_\beta = 1$ throughout the reminder of this paper, without loss of generality.

\subsection{Sparsity and Generalization}

\begin{figure}
\centering
\begin{minipage}{0.495\textwidth}
\centering
    \includegraphics[width=2.5in]{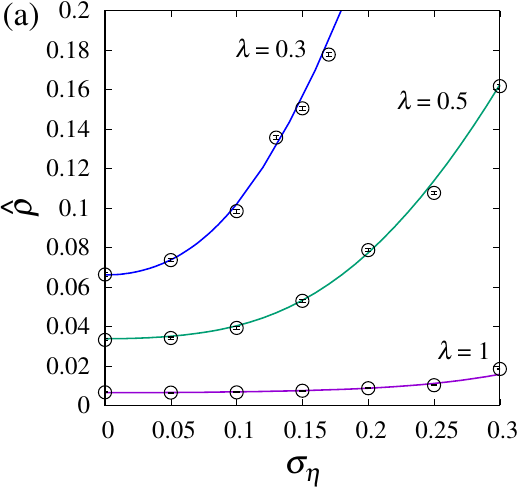}
    \end{minipage}
    \begin{minipage}{0.495\textwidth}
    \centering
    \includegraphics[width=3.5in]{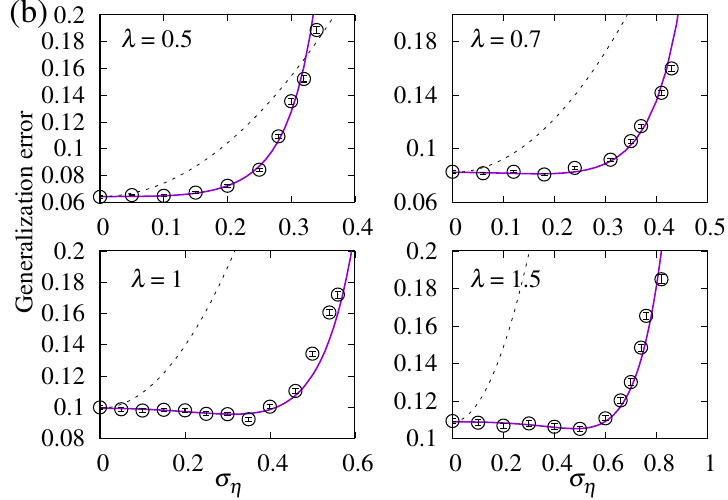}
    \end{minipage}
    \caption{
$\sigma_\eta$-dependence of (a) the fraction of nonzero components in the estimator and (b) the generalization error for various regularization parameters $\lambda$ under objective perturbation. Circles ($\circ$) denote AMP results and solid lines show the corresponding analysis by replica method. 
In the AMP experiments, we set $p=1000$, $\alpha=0.5$, $\rho=0.1$, and $\sigma_\xi=0.1$. AMP results are averaged over 100 datasets; error bars indicate the standard error. For each dataset, a single realization of privacy noise $\bm{\eta}$ is generated. The dashed lines in (b) shows the generalization error under output perturbation.}
\label{fig:vs_sigma_eta}
\end{figure}

\begin{figure}
\begin{minipage}{0.495\hsize}
\centering
    \includegraphics[width=2.5in]{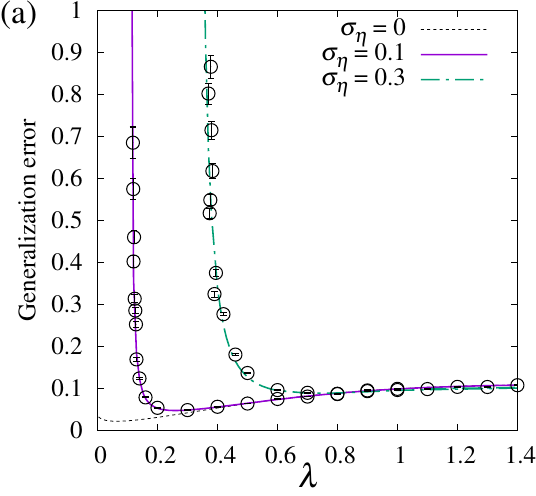}
    \end{minipage}
    \begin{minipage}{0.495\hsize}
    \centering
    \includegraphics[width=2.5in]{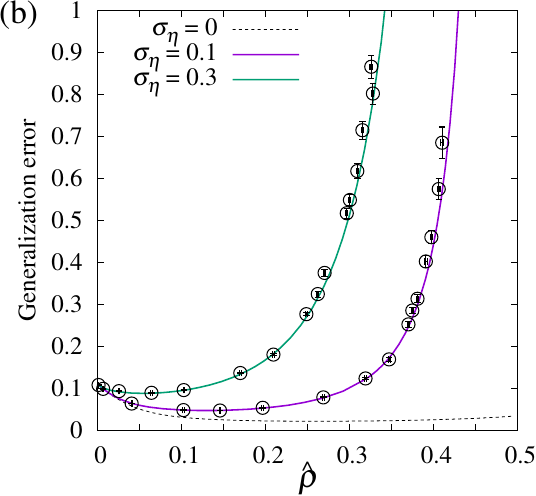}    
    \end{minipage}
    \caption{
(a) $\lambda$-dependence of the generalization error for different privacy noise levels ($\sigma_\eta=0.1$, $0.3$) compared with the noiseless case. 
(b) Generalization error as a function of the resulting $\widehat{\rho}$. 
The experimental setup and plotting style are the same as in Fig.~\ref{fig:vs_sigma_eta}.}
\label{fig:Gen_err_Gauss_replica}
\end{figure}

Fig.~\ref{fig:vs_sigma_eta}(a) shows the fraction of nonzero components in the AMP estimator, denoted by $\widehat{\rho}=\frac{1}{p}\|\widehat{\bm{\beta}}({\cal D},\bm{\eta})\|_0$, as a function of the privacy noise level $\sigma_\eta$.
As $\sigma_\eta$ increases, the estimator becomes progressively denser. This effect is more pronounced for smaller values of the regularization parameter $\lambda$. Hence, weaker regularization combined with privacy noise promotes denser solutions.

Figure~\ref{fig:vs_sigma_eta}(b) presents the corresponding generalization error as a function of $\sigma_\eta$.
For certain values of $\lambda$, the error exhibits a non-monotonic behavior: it initially decreases with increasing $\sigma_\eta$ and then increases. In particular, when $\lambda$ is sufficiently large, such as $\lambda=1$ and $\lambda=1.5$, a finite level of privacy noise yields a lower generalization error than in the noiseless case.
For comparison, we also show the generalization error under output perturbation in Fig.~\ref{fig:vs_sigma_eta}(b) (dashed lines). 
Under Setting~\textbf{D}, the generalization error under output perturbation, eq. \eqref{eq:gen_output}, is given by 
\begin{align}
E_{\mathrm{gen}}(\bm{\eta}) = E_{\mathrm{gen}}(\bm{0}) + \sigma_\eta^2,
\end{align}
which increases monotonically with the noise level.

The non-monotonic behavior of generalization error under objective perturbation is closely related to the densification observed in Fig.~\ref{fig:vs_sigma_eta}(a). 
For large $\lambda$, the estimator is overly sparse without privacy noise. 
Increasing $\sigma_\eta$ modifies the effective thresholding condition, leading to the activation of additional components and compensating for the shrinkage induced by strong regularization, thereby improving predictive performance.

Fig.~\ref{fig:Gen_err_Gauss_replica}(a) shows the $\lambda$-dependence of the generalization error under objective preturbation at $\alpha=0.5$, $\rho=0.1$, and $\sigma_\xi=0.1$.
Consistent with Fig.~\ref{fig:vs_sigma_eta}(b), increasing the privacy noise generally increases the generalization error, and for sufficiently small $\lambda$ the error can become very large.
Moreover, as the privacy noise level increases, the divergence of the error occurs at larger values of $\lambda$. 
This indicates that models with weaker regularization are more sensitive to objective noise in terms of generalization performance.
This behavior can be attributed to the fact that weakly regularized models have greater flexibility, which allows them to fit the privacy noise in addition to the true signal, thereby increasing the generalization error.

Figure~\ref{fig:Gen_err_Gauss_replica}(b) shows the same results as a function of the non-zero component in the estimator, $\widehat{\rho}$.
Even when $\widehat{\rho}$ is the same, larger privacy noise leads to a higher generalization error.
This indicates that $\widehat{\rho}$ alone is insufficient to characterize the prediction performance. 
Even when the sparsity level $\widehat{\rho}$ is the same, different choices of parameters (such as $\lambda$, $\alpha$, $\rho$) can lead to different generalization errors.

To analyze these behaviors more systematically, we study the estimator using the state evolution associated with the AMP algorithm. 
This analysis characterizes the macroscopic quantities governing the estimator and, in particular, reveals a proportional relationship between the training and generalization errors.

\subsection{Asymptotic behavior based on State Evolution}

\subsubsection{AMP's decoupling and state evolution} 

The dependence of the AMP trajectory on the dataset and privacy noise appears through 
$\{\mathrm{m}_1,\ldots,\mathrm{m}_p\}$ and $\Sigma$, 
which are random variables determined by the dataset and privacy noise. 
Under Setting~\textbf{D}, the variance of $\Sigma$ is sufficiently small and negligible in the high-dimensional limit.
By analyzing the statistical properties of $\{\mathrm{m}_1,\ldots,\mathrm{m}_p\}$, we obtain the following result.
\begin{Thm}[AMP decoupling with privacy noise]
\label{Thm:DE}
Consider a component of the AMP estimator at step $t$, 
$\widehat{\bm{\beta}}_{\mathrm{AMP}}^{(t)}$, with assigned privacy noise $\eta$ 
and ground truth $\beta^{(0)}$. 
Denote the component by 
$\widehat{\beta}^{(t)}_{\mathrm{AMP}}(X,\beta^{(0)},\eta)$.
If the privacy noise satisfies Assumption~P, then in the high-dimensional limit 
$n,p\to\infty$ with $\alpha=n/p=O(1)$, 
the estimator is asymptotically equivalent in distribution to
\begin{align}
    \widehat{\beta}^{(t+1)}_{(\mathrm{AMP})}(X|\beta^{(0)},\eta)\overset{d}{=}\mathbb{M}\left(\Sigma^{(t)},\beta^{(0)}+\sigma_z^{(t)}z-\eta\Sigma^{(t)}\right)
\end{align}
where $\sigma_z^{(t)}=\sqrt{E^{(t)}/\alpha}$, 
$\Sigma^{(t)}=(1+V^{(t)})/\alpha$, and $z\sim{\cal N}(0,1)$. 
The parameters $E^{(t)}$ and $V^{(t)}$ satisfy
\begin{align}
    E^{(t)}&=\int d\beta^{(0)}\phi_0(\beta^{(0)}) \int d\eta P_\eta(\eta)\int Dz \left(\beta^{(0)}-\mathbb{M}\left(\Sigma^{(t)},\beta^{(0)}+\sigma_zz-\eta\Sigma^{(t)}\right)\right)^2+\sigma_\xi^2\\
    V^{(t)}&=\int d\beta^{(0)}\phi_\beta(\beta^{(0)}) \int d\eta P_\eta(\eta)\int Dz\mathbb{V}(\Sigma^{(t)},\beta^{(0)}+\sigma_zz-\eta\Sigma^{(t)}).
\end{align}
\end{Thm}
Theorem \ref{Thm:DE} states that, in the high-dimensional limit, the joint dynamics of the AMP iterates become asymptotically equivalent to a collection of independent scalar estimation problems driven by an effective Gaussian noise \citep{bayati2011dynamics}. 
This characterization is known as \emph{state evolution}.
More precisely, in the presence of privacy noise, the injected objective noise decomposes into two qualitatively distinct components: 
(i) a local contribution through the coordinate-wise noise, and 
(ii) a non-local contribution arising from the remaining coordinates, which affects each coordinate indirectly through the macroscopic quantities $E^{(t)}$ and $V^{(t)}$. 
As a result, the AMP trajectory becomes asymptotically decoupled across coordinates, and the dynamics can be described by a scalar stochastic recursion that depends on each $\eta_i$ at the coordinate level and on $E^{(t)}$ and $V^{(t)}$ at the macroscopic level. 
The derivation of Theorem~\ref{Thm:DE} is provided in Supplement Material.

The state evolution parameters have the following meanings:
\begin{align}
E^{(t)} &= \frac{1}{p}\|\widehat{\bm{\beta}}_{\mathrm{AMP}}^{(t)}({\cal D,\bm{\eta}})-\bm{\beta}^{(0)}\|^2 + \sigma_\xi^2,\\
V^{(t)} &= \lim_{\tau\to 0}\frac{\tau}{p}\sum_{i=1}^{p}\mathrm{Var}^{(t)}_\tau(\beta),
\end{align}
at sufficiently large $p$,
where $\mathrm{Var}^{(t)}_\tau(\cdot)$ denotes the posterior variance at $\tau$ as defined in \eqref{eq:def_posterior} at step $t$.
The state evolution parameter $E$ corresponds to the generalization error under Setting $\mathbf{D}$, and $V$ corresponds to the local variance of the estimator.
Although the quantities on the right-hand sides are defined for a fixed realization of the dataset ${\cal D}$ and the objective noise $\bm{\eta}$, 
in the high-dimensional limit $n,p\to\infty$ with $\alpha=n/p=O(1)$, the empirical generalization error concentrates around its expectation with respect to the randomness of ${\cal D}$ and $\bm{\eta}$. 
Consequently, a typical realization of the AMP estimator yields the same asymptotic error as that predicted by state evolution.
In Figs.~\ref{fig:vs_sigma_eta} and \ref{fig:Gen_err_Gauss_replica}, we compare the AMP results under a fixed realization of the objective privacy noise with the corresponding result obtained from state evolution and analysis by replica method, which characterize the average over $\bm{\eta}$ in the high-dimensional limit (see Supplement).
We observe good agreement between them, indicating that the results do not depend on the specific realization of $\bm{\eta}$.

The presence of privacy noise does not alter the overall structure of the state evolution equations. 
Consequently, many arguments developed for the noiseless setting carry over directly to the present setting with privacy noise.

\subsubsection{Proportionality between training and generalization errors}

Utilizing the state evolution form, we can show a relationship 
which holds without depending on the existence of the objective privacy noise.
\begin{claim}[Proportionality between two errors]
\label{claim:err_prop}
Under the RS assumption, for arbitrary values of $\sigma_\xi$, $\sigma_\eta$, $\alpha$, $\rho$, and $\lambda$, the generalization and training errors computed at the fixed point of AMP, denoted by $E_{\mathrm{gen}}^{(\mathrm{AMP})}$ and $E_{\mathrm{train}}^{(\mathrm{AMP})}$, satisfy the following proportionality in the high-dimensional limit:
\begin{align}
    \frac{E_{\mathrm{gen}}^{(\mathrm{AMP})}}{E_{\mathrm{train}}^{(\mathrm{AMP})}} = \left(1+V\right)^2.
    \label{eq:proportionality}
\end{align}
\end{claim}
The detail is shown in Supplement Material.

In Fig.~\ref{fig:Gen_err_Gauss_replica}(b), we show the ratio between the generalization error and the training error obtained from the AMP algorithm, together with the corresponding value predicted by the state evolution (SE) equations in the large-$p$ limit, which is given by the replica method. 
This proportionality reveals that the effect of privacy noise enters both errors through a common multiplicative factor. 
In particular, the gap between training and generalization errors is fully controlled by the scalar parameter $V$.

This behavior contrasts with output perturbation. 
Under Setting~\textbf{D}, output perturbation satisfies
$E_{\mathrm{gen}}(\bm{\eta}) - E_{\mathrm{train}}(\bm{\eta})
= 
E_{\mathrm{gen}}(\bm{0}) - E_{\mathrm{train}}(\bm{0})$,
indicating that the generalization gap is unaffected by the noise. 
In contrast, objective perturbation modifies this gap through the factor $(1+V)^2$, 
demonstrating a qualitative difference in how the two mechanisms influence errors.

\begin{figure}
\begin{minipage}{0.495\textwidth}
    \centering
    \includegraphics[width=2.5in]{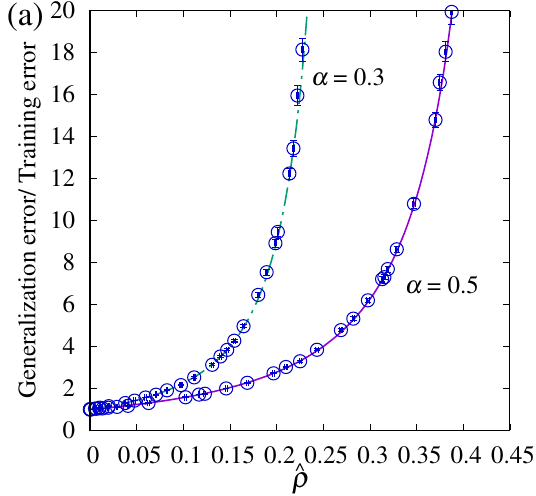}
    \end{minipage}
    \begin{minipage}{0.495\textwidth}
    \centering
    \includegraphics[width=2.5in]{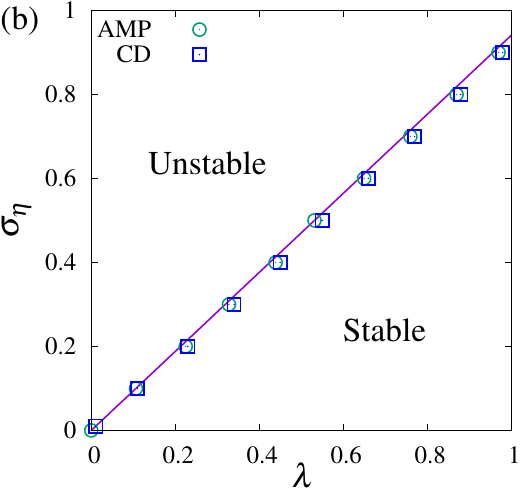}
    \end{minipage}
    \caption{(a) Dependence of the ratio between generalization and training errors on $\widehat{\rho}$ at $\rho=0.1$, $\sigma_\xi=1$, and $\sigma_\eta=0.1$, for $\alpha=0.5$ and $\alpha=0.3$, as predicted by state evolution in the high-dimensional limit. Circles represent the AMP results.
(b) Parameter region where AMP does not converge at $\alpha=0.5$, $\rho=0.5$ and $\sigma_\xi=0.1$. The solid line denotes the boundary predicted by the replica method, circles and squares indicate the empirical convergence limit of AMP, and squares indicate that of coordinate descent. For both AMP and coordinate descent, convergence was evaluated over 100 independent runs, and a point is marked as non-convergent if at least one run fails to converge.}
    \label{fig:prediction_training_ratio}
\end{figure}

\subsubsection{Convergence condition of AMP}

The AMP algorithm does not always converge. The stability of the AMP algorithm can be characterized by its local stability condition associated with state evolution.
In the present setting, the stability condition is given by \citep{krzakala2012probabilistic,sakata2023prediction}
\begin{align}
    \frac{\widehat{\rho}}{\alpha}<1,
    \label{eq:AT}
\end{align}
where $\widehat{\rho}$ denotes the asymptotic fraction of nonzero components in the estimator.
This condition simply requires that the number of active coefficients does not exceed the number of observations.
The results presented in this paper are valid only when \eqref{eq:AT} is satisfied.

Figure~\ref{fig:prediction_training_ratio}(b) illustrates the parameter region where this stability condition is satisfied. 
The boundary of the stable region depends primarily on the ratio $\alpha$, with weaker dependence on other parameters. 
Although the convergence condition is derived from AMP, it is also related to the behavior of coordinate descent algorithms, as shown in Fig.\ref{fig:prediction_training_ratio}(b).
In prior work without privacy noise, non-unique solution regions of coordinate descent coincide with AMP divergence regions \citep{sakata2018approximate,obuchi2019cross}. We find that this correspondence holds under privacy noise as well.

Although the optimization problem with objective perturbation remains convex, the above condition shows that the estimator can enter a regime where AMP fails to converge. 
This behavior is closely related to the sparsity of the solution. 
In particular, objective perturbation tends to increase the number of active coefficients, making the estimator denser. 
As a result, the condition $\widehat{\rho}/\alpha<1$ is more easily violated compared to the noiseless case.

In the high-dimensional limit $p,n\to\infty$ with $\alpha = n/p = O(1)$, combining the state evolution analysis with an analysis by replica method gives the following expression for the state evolution parameter:
\begin{align}
    V=\frac{\widehat{\rho}}{\alpha-\widehat{\rho}}.
    \label{eq:chi_expression}
\end{align}
Equation~\eqref{eq:chi_expression} shows that $V$ diverges as $\alpha$ approaches $\widehat{\rho}$, i.e., when condition~\eqref{eq:AT} is violated. This divergence explains the observed divergence of the empirical ratio in Fig.~\ref{fig:Gen_err_Gauss_replica}(b).

\section{Asymptotic Distribution of the AMP Estimator}

\begin{figure}
    \centering
    \includegraphics[width=\textwidth]{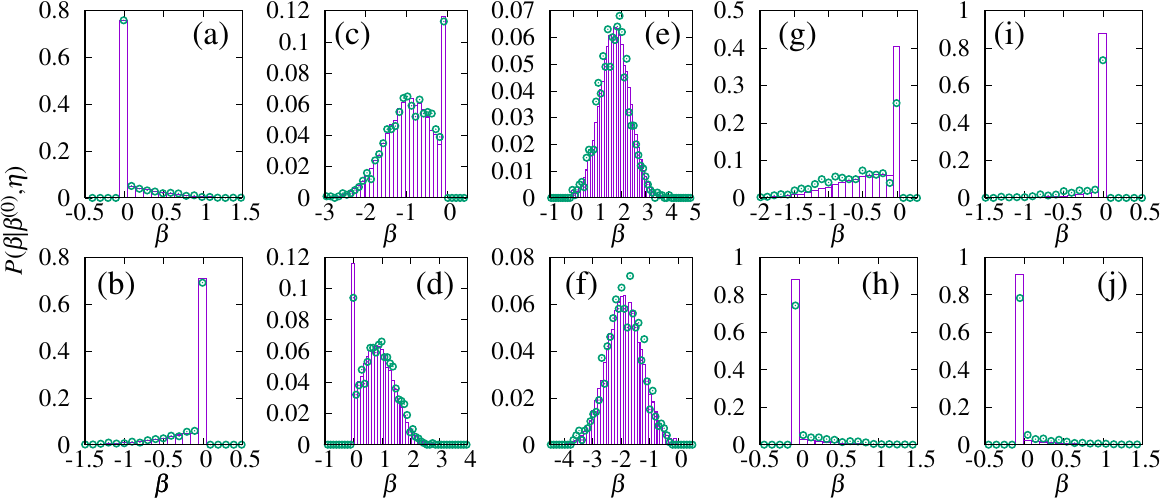}
    \caption{Comparison of distribution of the estimators with respect to data by AMP ($\circ$) and DE (boxes) at $\alpha=0.5$, $\rho=0.3$, $\sigma_\xi=0.1$, $\lambda=0.5$ and $\sigma_\eta=0.1$. Distributions for different $\beta^{(0)}$ and $\eta$ are shown: (a) $\beta^{(0)}=1.0629,~\eta = 0.0460$, (b) $\beta^{(0)}=-1.0412,~ \eta = -0.0238$, (c) $\beta^{(0)}=-2.1796,~\eta = -0.0079$, (d) $\beta^{(0)}=2.1290,~\eta = -0.0317$, (e) $\beta^{(0)}=3.2188,~\eta = 0.0556$, (f) $\beta=-3.2333,~\eta = -0.0321$, (g) $\beta^{(0)}=-1.0095,~\eta = 0.2907$, (h) $\beta^{(0)}=0.1017,~\eta = -0.3261$, (i) $\beta^{(0)}=0,~\eta = 0.3579$, (j) $\beta^{(0)} = 0, \eta = -0.3233$. }
    \label{fig:beta_dist_fixed_eta}
\end{figure}

\subsection{Equivalent Distribution over the Data (Fixed Privacy Noise)}

To evaluate the typical information-theoretic privacy criterion, the on-average KL divergence, we examine the component-wise distribution of estimates produced by AMP. According to Theorem~\ref{Thm:DE}, it suffices to consider the conditional distribution of a single component given fixed values of $\beta^{(0)}$ and $\eta$.
We thus introduce the following conditional probability density function
for a sufficiently small increment $d\beta \ll 1$:
\begin{align}
\mathbb{P}^{\mathrm{AMP}}_{|\beta^{(0)},\eta}(\beta|\beta^{(0)},\eta)d\beta
:=\int dX d\bm{\xi}\, P_X(X)P_\xi(\bm{\xi})
\mathbb{I}(\beta\leq\widehat{\beta}_{\mathrm{AMP}}({\cal D},\eta)<\beta+d\beta).
\label{eq:def_P_AMP}
\end{align}
Following Theorem~\ref{Thm:DE}, the conditional distribution of the AMP estimate
given $(\beta^{(0)},\eta)$ converges in probability to the distribution induced
by the state evolution:
\begin{align}
\mathbb{P}^{\mathrm{AMP}}_{|\beta^{(0)},\eta}(\beta|\beta^{(0)},\eta)
\xrightarrow{\mathbb{P}}
\mathbb{P}^{\mathrm{SE}}_{|\beta^{(0)},\eta}(\beta|\beta^{(0)},\eta),
\label{eq:equivalence_dist_under_noise}
\end{align}
where the limiting distribution $\mathbb{P}^{\mathrm{SE}}_{|\beta^{(0)},\eta}$
is defined by
\begin{align}
        \mathbb{P}_{|\beta^{(0)},\eta}^{\mathrm{DE}}(\beta|\beta^{(0)},\eta)d\beta&=\int Dz\mathbb{I}\left(\beta\leq\mathbb{M}\left(\frac{1+s_\theta}{\alpha},\beta^{(0)}+\sigma_zz-\frac{\eta(1+s_\theta)}{\alpha}\right)<\beta+d\beta\right).\label{eq:def_P_DE}
\end{align}
Fig.~\ref{fig:beta_dist_fixed_eta} compares the distributions obtained from AMP,
$\mathbb{P}^{\mathrm{AMP}}_{|\beta^{(0)},\eta}$ at $p=5000$, and from state evolution,
$\mathbb{P}^{\mathrm{SE}}_{|\beta^{(0)},\eta}$, for fixed realizations of $\bm{\eta}$ and $\bm{\beta}^{(0)}$.
For AMP, the distribution is estimated using the fixed point over 1000 realizations of $X$ and $\bm{\xi}$.
In both AMP and SE, the distributions
are computed numerically on a discretized grid with step size $\Delta\beta = 10^{-2}$.
Following the decoupling principle, we compare the distributions of components
associated with given values of $\beta^{(0)}$ and $\eta$, for various choices of
these parameters.

The AMP results qualitatively reproduce the distributions predicted by SE,
although the accuracy depends on the magnitudes of $\beta^{(0)}$ and $\eta$.
As shown in Fig.~\ref{fig:beta_dist_fixed_eta}(a)--(f), where $|\eta|$ is relatively
small (up to $O(10^{-2})$), AMP agrees well with SE. Minor discrepancies can be
attributed to finite-size effects, the limited number of samples, and discretization
errors. When $\beta^{(0)}$ is small, the agreement becomes even better, since the
distribution is dominated by a single peak with small residual tails.
In contrast, as shown in panels (g)--(j), for larger $|\eta|$ on the order of
$10^{-1}$, noticeable deviations appear even for small $\beta^{(0)}$, particularly
in the peak amplitude. However, the overall qualitative behavior remains consistent with the SE prediction, and the agreement is expected to improve with larger system sizes and an increased number of samples used to estimate the distribution.

\subsection{Equivalent Distribution over the privacy noise (Fixed data)}

Since \eqref{eq:equivalence_dist_under_noise} holds for any fixed values of
$\beta^{(0)}$ and $\eta$, averaging over the privacy noise yields, for any $\beta$,
\begin{align}
\int d\eta~
\mathbb{P}^{\mathrm{AMP}}_{|\beta^{(0)},\eta}(\beta|\beta^{(0)},\eta)P_\eta(\eta)
=
\int d\eta~
\mathbb{P}^{\mathrm{SE}}_{|\beta^{(0)},\eta}(\beta|\beta^{(0)},\eta)P_\eta(\eta).
\label{eq:equiv_P_average}
\end{align}
Using the definitions in \eqref{eq:def_P_AMP} and \eqref{eq:def_P_DE},
this identity can be rewritten as
\begin{align}
\int dX d\bm{\xi}\,P_X(X)P_\xi(\bm{\xi})
\mathbb{P}^{\mathrm{AMP}}_{|\beta^{(0)},X}(\beta|\beta^{(0)},X)
=
\int Dz~
\mathbb{P}^{\mathrm{SE}}_{|\beta^{(0)},z}(\beta|\beta^{(0)},\sigma_z z),
\label{eq:equivalence_dist_change}
\end{align}
where the conditional distributions are defined as
\begin{align}
    \mathbb{P}_{|\beta^{(0)},X}^{\mathrm{AMP}}(\beta|\beta^{(0)},X)d\beta&=\int d\eta P_\eta(\eta) \mathbb{I}(\beta\leq\widehat{\beta}_{\mathrm{AMP}}({\cal D},\eta)<\beta+d\beta)\\
    \mathbb{P}_{|\beta^{(0)},z}^{\mathrm{DE}}(\beta|\beta^{(0)},\sigma_zz)d\beta&=\int d\eta P_\eta(\eta) \mathbb{I}\left(\beta\leq\mathbb{M}\left(\frac{1+s_\theta}{\alpha},\beta^{(0)}+\sigma_zz-\frac{\eta(1+s_\theta)}{\alpha}\right)<\beta+d\beta\right).
\end{align}
Accordingly, we obtain the following equivalence in distribution.
\begin{claim}[Equivalent distribution induced by objective privacy noise]

For a fixed dataset ${\cal D}$, the distribution of a component of the
AMP estimator over the privacy noise is asymptotically equivalent to the
scalar distribution generated by the state evolution with Gaussian noise $z$:
\begin{align}
\mathbb{P}_{|\beta^{(0)},X}^{\mathrm{AMP}}(\beta|\beta^{(0)},X)
\overset{d}{=}
\mathbb{P}_{|\beta^{(0)},z}^{\mathrm{SE}}(\beta|\beta^{(0)},\sigma_z z).
\label{eq:equivalence_dist_under_data}
\end{align}

\end{claim}

Moreover, \eqref{eq:equivalence_dist_change} implies that, for any measurable
function $a(\beta)$,
\begin{align}
    \int d\beta~a(\beta)\mathbb{P}^{\mathrm{AMP}}_{|\beta^{(0)},X}(\beta|\beta^{(0)},X)\overset{d}{=}\int d\beta~a(\beta)\mathbb{P}^{\mathrm{SE}}_{|\beta^{(0)},z}(\beta|\beta^{(0)},\sigma_zz).
    \label{eq:equivalence_mean}
\end{align}
The equivalent scalar distribution
$\mathbb{P}_{|z,\beta^{(0)}}^{\mathrm{SE}}(\beta|\beta^{(0)},\sigma_z z)$
admits a closed-form expression, derived in the Supplementary Material:
\begin{align}
\nonumber
    \mathbb{P}_{|\beta^{(0)},z}^{\mathrm{SE}}(\beta|\beta^{(0)},\sigma_zz)&=(1-\widehat{r}^{\mathrm{SE}}(\beta^{(0)}+\sigma_zz,\Sigma))\delta(\beta)\\
    &\hspace{1.0cm}+\frac{1}{\sqrt{2\pi}\Sigma\sigma_\eta}\exp\left(-\frac{\left(\beta-(\beta^{(0)}+\sigma_zz)+\mathrm{sgn}(\beta)\lambda\Sigma\right)^2}{2\Sigma^2\sigma_\eta^2}\right),
    \label{eq:dist_beta_under_z}
\end{align}
where 
\begin{align}
    \widehat{r}^{\mathrm{SE}}(\beta^{(0)}+\sigma_zz,\Sigma)=\frac{1}{2}\left\{\mathrm{erfc}\left(\frac{-(\beta^{(0)}+\sigma_zz)+\lambda\Sigma}{\sqrt{2}\Sigma\sigma_\eta}\right)+\mathrm{erfc}\left(\frac{\beta^{(0)}+\sigma_zz+\lambda\Sigma}{\sqrt{2}\Sigma\sigma_\eta}\right)\right\}.
\end{align}
This quantity represents the probability, with respect to the objective noise,
that a component with true signal $\beta^{(0)}$ is estimated as nonzero under a fixed realization of $z$.
This interpretation follows directly from \eqref{eq:equivalence_mean} by setting
$a(\beta)=\mathbb{I}(|\beta|>0)$, which yields
\begin{align}
\widehat{r}^{\mathrm{AMP}}({\cal D})
:=
\int d\beta\,\mathbb{I}(|\beta|>0)\mathbb{P}^{\mathrm{AMP}}(\beta|{\cal D})
\overset{d}{=}
\widehat{r}^{\mathrm{SE}}(\beta^{(0)}+\sigma_z z,\Sigma).
\end{align}
In the limit $\sigma_\eta \to 0$,
$\widehat{r}^{\mathrm{SE}}(\beta^{(0)},\sigma_z z)$ reduces to
$\Theta(|\beta^{(0)}+\sigma_z z|>\lambda\Sigma)$,
and the distribution
$\mathbb{P}_{|\beta^{(0)},z}^{\mathrm{SE}}$
collapses to a Dirac delta located at the soft-thresholding estimator,
thereby recovering the original LASSO solution.

\begin{figure}
    \centering
    \includegraphics[width=\textwidth]{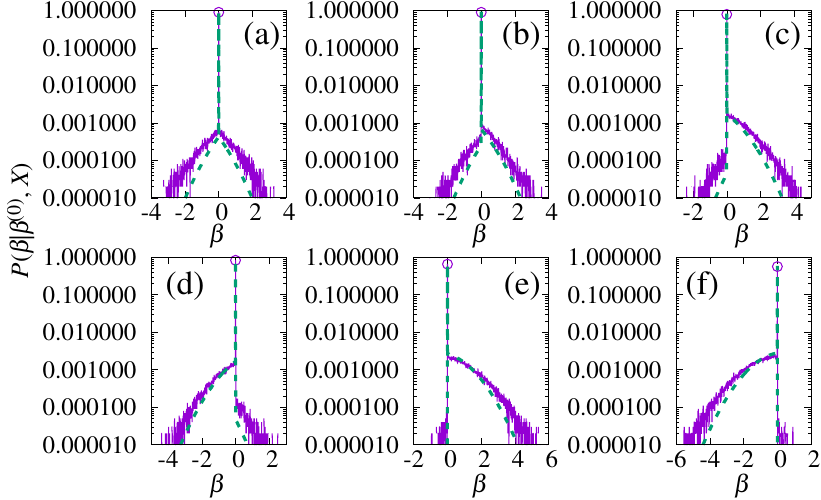}
    \caption{
    Examples of the distributions of the AMP estimates induced by the injected privacy noise (solid lines) and the corresponding SE predictions (dashed lines) at $\alpha=0.5$, $\rho=0.5$, $\sigma_y=0.1$, $\lambda=1$, and $\sigma_\eta=0.5$. For AMP, we set $N=500$, and 1000 realizations of the objective noise are used in the computation. Panels (a)-(f) correspond to $\beta^{(0)}=0$, $0.3554$, $1.308$, $-1.279$, $2.191$, and $-2.432$, respectively. The circle at $\beta=0$ indicates the height of the peak of the AMP distribution.}
    \label{fig:beta_dist_fixed_data}
\end{figure}

Fig.~\ref{fig:beta_dist_fixed_data} shows the distribution of the AMP estimate
with respect to the privacy noise for a fixed dataset. For comparison, the
distribution predicted by state evolution is shown by dashed lines.
We present results for several values of $\beta^{(0)}$, where the distributions
are averaged over 100 realizations of the dataset (for AMP) or of $z$ (for SE),
since establishing a one-to-one correspondence between $X$ and $z$ is non-trivial
in the present framework.
Overall, we observe good agreement between AMP and SE. 
In the nonzero case, the mass of the
distribution is biased toward the sign of $\beta^{(0)}$, reflecting the effect
of the underlying signal.

\section{Asymptotic Evaluation of the On-Average KL Divergence}

AMP does not provide the full joint distribution of all variables on a non-tree factor graph.
We therefore focus on comparing marginal distributions under one-point-mutant datasets.
Starting from the original definition of the OnAveKL in \eqref{eq:def_OnAveKL},
we introduce the component-wise On-Average KL divergence, defined as
\begin{align}
    \mathrm{cwOnAveKL}=\frac{1}{n}\sum_{\mu=1}^n\mathbb{E}_{{\cal D},{\cal D}_\mu^\prime}\left[\int d\beta_i\prod_{i=1}^Np_i(\beta_i|{\cal D})\sum_{i=1}^N\ln\frac{p_i(\beta_i|{\cal D})}{p_i(\beta_i|{\cal D}_\mu^\prime)}\right],
\end{align}
where $p_i(\beta_i|{\cal D})$ denotes the marginal posterior distribution of the $i$-th component under dataset ${\cal D}$.

To provide a reference scale for interpreting the magnitude of cwOnAveKL, consider the following simple Gaussian example.
When the two distributions over the $p$ components are i.i.d.\ Gaussian with identical variance $\sigma^2$ but different means, and the mean difference is uniformly $\mu$ across all components, the divergence reduces to
\begin{align}
    \mathrm{cwOnAveKL}=\frac{p\mu^2}{2\sigma^2}.
\end{align}
In particular, the condition $\mathrm{OnAveKL}=\varepsilon/\sigma^2$ corresponds,
in the Gaussian case, to a component-wise mean deviation of order
\begin{align}
    \mu=\sqrt{\frac{2\varepsilon}{p}}.
\end{align}
Although the distributions considered in this work are not always Gaussian, in particular for objective perturbation case,
this relation provides a useful reference scale for interpreting the magnitude of cwOnAveKL.

\subsection{cwOnAveKL under Output Perturbation}

First, we evaluate the cwOnAveKL for output perturbation, where the calculation simplifies significantly in the infinite dimensional case. 
In this regime, support changes induced by a one-point perturbation occur only rarely. Namely, components that are zero under $\mathcal{D}$ typically remain zero under $\mathcal{D}_\mu^\prime$, 
and their contribution to the KL divergence after adding output noise is negligible, as their distributions are the same.
Therefore, only the active components contribute to cwOnAveKL.

For the non-zero components, output perturbation induces Gaussian distributions
${\cal N}(\widehat{\beta}_i(\mathcal{D},\bm{0}),\sigma_\eta^2)$ and
${\cal N}(\widehat{\beta}_i(\mathcal{D}_\mu^\prime,\bm{0}),\sigma_\eta^2)$.
Since these two distributions share the same variance $\sigma_\eta^2$, their KL divergence is given by the squared difference of the means.
Therefore, we obtain
\begin{align}
    \mathrm{cwOnAveKL}=\frac{1}{n\sigma_\eta^2}\sum_{\mu=1}^n\mathbb{E}_{{\cal D},{\cal D}^\prime_\mu}\left[\sum_{i=1}^p\mathbb{I}(|\widehat{\beta}_i({\cal D},\bm{0})|>0)(\widehat{\beta}_i({\cal D},\bm{0})-\widehat{\beta}_i({\cal D}_\mu^\prime,\bm{0}))^2\right].
\end{align}

Using the AMP expression, combined with replica expression, explained by Supplement Material, $\textrm{cwOnAveKL}$ for output perturbation is given by
\begin{align}
    \mathrm{cwOnAveKL}=\frac{E_0\widehat{\rho}_0}{\alpha^2\sigma_\eta^2},
    \label{eq:cwOnAveKL_out}
\end{align}
where $E_0=E(\bm{\eta}=0)$ and $\widehat{\rho}_0=\frac{1}{p}\mathbb{E}_{{\cal D}}[\|\widehat{\bm{\beta}}({\cal D},\bm{0})\|_0]$.

\subsection{cwOnAveKL under objective perturbation}

For computing cwOnAveKL for objective perturbation, we need to compare two marginal distributions under ${\cal D}$ and ${\cal D}_\mu^\prime$.
We represent the original dataset ${\cal D}$ and
its one-point-mutant counterpart ${\cal D}_\mu^\prime$ as
${\cal D}=\{{\cal D}_{\setminus\mu},\bm{d}_\mu\}$ and
${\cal D}_\mu^\prime=\{{\cal D}_{\setminus\mu},\bm{d}_\mu^\prime\}$, respectively,
where ${\cal D}_{\setminus\mu}$ denotes the LOO dataset obtained by removing the $\mu$-th sample from ${\cal D}$.
We make use of the following claim, which shows that cwOnAveKL can be expressed solely in terms of the decoupled distributions.

\begin{claim}[Distribution of estimate under One-Point-Mutant data]
\label{claim:OnePointMutant}
Let us consider a functional ${\cal A}({\cal D}, {\cal D}_{\mu}^\prime)=a(\mathbb{P}_{|\beta^{(0)},X}^{\mathrm{AMP}}(\beta|,\beta^{(0)},\{{\cal D}_{\setminus\mu},\bm{d}_\mu\}),\mathbb{P}^{\mathrm{AMP}}_{|\beta^{(0)},X^\prime_\mu}(\beta|\{{\cal D}_{\setminus\mu},\bm{d}_\mu^\prime\}))$ defined for an arbitrary function $a(\cdot,\cdot)$. Then, for any such $a(\cdot,\cdot)$, the following equality in distribution holds:
\begin{align}
    {\cal A}({\cal D},{\cal D}_\mu^\prime)&\overset{d}{=}
    a\left(\mathbb{P}_{|\beta^{(0)},z}^{\mathrm{SE}}(\beta|\beta^{(0)},\sigma_z^{\setminus\Delta}z+\sqrt{\Delta}\zeta),\mathbb{P}^{\mathrm{SE}}_{|\beta^{(0)},z}(\beta|\beta^{(0)},\sigma_z^{\setminus\Delta}z+\sqrt{\Delta}\zeta^\prime)\right)
\end{align}
where $\sigma_z^{\setminus\Delta}=\sqrt{\sigma_z^2-\Delta}$,
$\Delta=E/(\alpha n)$,
and $z$, $\zeta$ and $\zeta^\prime$ are independent standard Gaussian random variables.
\end{claim}
The derivation of Claim~\ref{claim:OnePointMutant} is provided in Supplement Material.
Here, $E$ denotes the generalization error for both the full dataset and the LOO dataset; we assume that their difference is negligible in the large-$n$ limit.

Utilizing Claim~\ref{claim:OnePointMutant} and choosing
$a(\mathbb{P},\mathbb{Q})=\mathbb{P}\ln(\mathbb{P}/\mathbb{Q})$,
the cwOnAveKL can be expressed as
\begin{align}
\nonumber
    &\mathrm{cwOnAveKL}\\
    &=\int Dz D\zeta D\zeta^{\prime}\sum_{i=1}^p\int d\beta_i\mathbb{P}_{|\beta^{(0)},z}^{\mathrm{SE}}(\beta_i|\beta_i^{(0)},\sigma_z^{\setminus\Delta}z+\Delta\zeta)\ln\frac{\mathbb{P}_{|z}^{\mathrm{SE}}(\beta_i|\beta_i^{(0)},\sigma_z^{\setminus\Delta}z+\Delta\zeta)}{\mathbb{P}_{|z}^{\mathrm{SE}}(\beta_i|\beta_i^{(0)},\sigma_z^{\setminus\Delta}z+\Delta\zeta^\prime)}.
    \label{eq:cwOnAveKL_mean}
\end{align}
Since $\Delta=O(n^{-1})$, we may expand \eqref{eq:cwOnAveKL_mean} to first order in
$\Delta$ for sufficiently large $n$, leading to the following result.
\begin{Coro}
The component-wise On-Average KL divergence for the AMP estimator is given by
\begin{align}
\mathrm{cwOnAveKL}&=\frac{ER}{\alpha^2\sigma_\eta^2},
\label{eq:cwOnAveKL_obj}
\end{align}    
where 
\begin{align}
    R=\sigma_\eta^2\int d\beta^{(0)}\phi_\beta(\beta^{(0)})\int Dz\left\{\frac{1}{1-\widehat{r}(\widehat{\mathrm{m}},\Sigma)}\left(\frac{\partial\widehat{r}(\widehat{\mathrm{m}},\Sigma)}{\partial\widehat{\mathrm{m}}}\right)^2+\frac{\partial^2\widehat{r}(\widehat{\mathrm{m}},\Sigma)}{\partial\widehat{\mathrm{m}}^2}+\frac{\widehat{r}(\widehat{\mathrm{m}},\Sigma)}{\Sigma^2\sigma_\eta^2}\right\},
    \label{eq:def_R}
    \end{align}
with $\widehat{\mathrm{m}}=\beta^{(0)}+\sigma_zz$. 

\end{Coro}
The derivation of \eqref{eq:def_R} is shown in Supplement Material. The term $R$ is the contribution of the sparisty to the privacy performance.

\subsection{Comparison of cwOnAveKLs}

\begin{figure}
    \begin{minipage}{0.32\textwidth}
    \centering
    \includegraphics[width=2in]{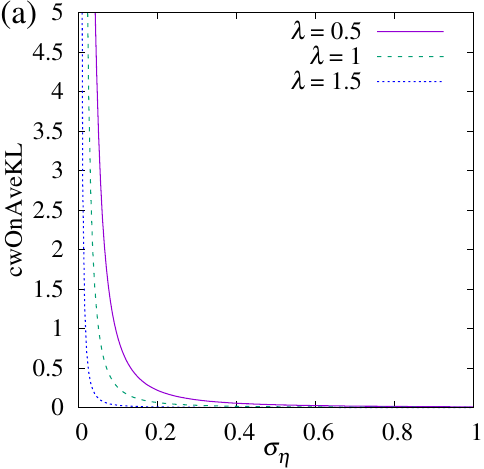}
    \end{minipage}
    \begin{minipage}{0.32\textwidth}
    \centering
    \includegraphics[width=2in]{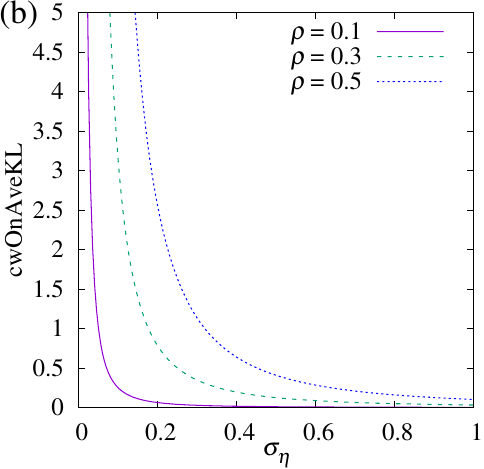}
    \end{minipage}
    \begin{minipage}{0.32\textwidth}
        \centering
    \includegraphics[width=2in]{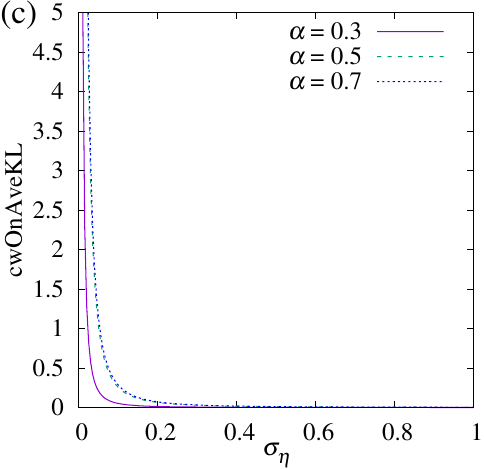}
    \end{minipage}
    \caption{cwOnAveKL under output perturbation as a function of the noise strength $\sigma_\eta$ for different parameter settings: (a) $\alpha=0.5$, $\rho=0.1$, (b) $\alpha=0.5$, $\lambda=1$, and (c) $\rho=0.1$, $\lambda=1$. The observation noise strength is set to $\sigma_\xi=0.1$.}
    \label{fig:cwOnAveKL_output}
\end{figure}

\begin{figure}
    \begin{minipage}{0.32\textwidth}
    \centering
    \includegraphics[width=2in]{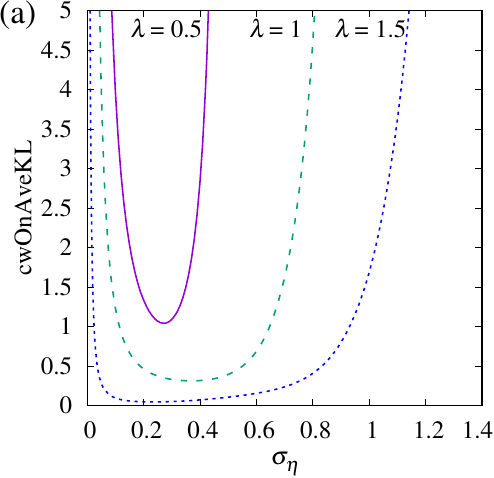}
    \end{minipage}
    \begin{minipage}{0.32\textwidth}
    \centering
    \includegraphics[width=2in]{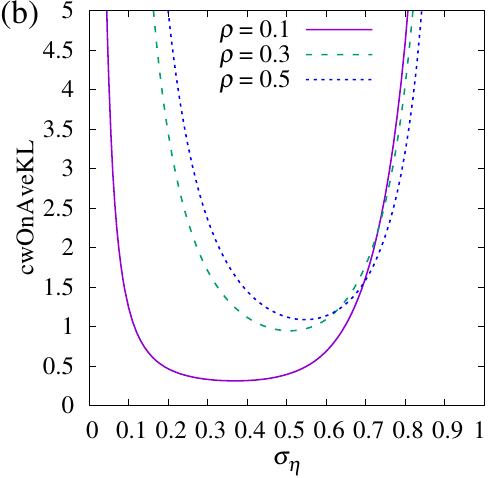}
    \end{minipage}
    \begin{minipage}{0.32\textwidth}
        \centering
    \includegraphics[width=2in]{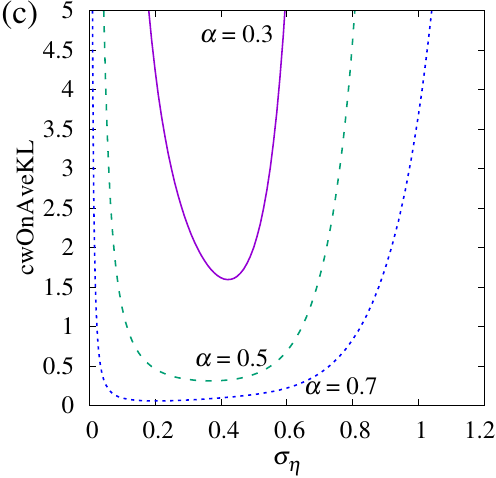}
    \end{minipage}
    \caption{cwOnAveKL under objective perturbation as a function of the noise strength $\sigma_\eta$ for different parameter settings: (a) $\alpha=0.5$, $\rho=0.1$, (b) $\alpha=0.5$, $\lambda=1$, and (c) $\rho=0.1$, $\lambda=1$. The observation noise strength is set to $\sigma_\xi=0.1$.}
    \label{fig:cwOnAveKL}
\end{figure}

The cwOnAveKLs for output and objective perturbations exhibit a similar structural feature: in both cases, the numerator is given by the generalization error multiplied by a sparsity-dependent coefficient (Eqs.~\eqref{eq:cwOnAveKL_out} and \eqref{eq:cwOnAveKL_obj}). This common structure highlights the role of sparsity in privacy.

We compare the cwOnAveKL for output and objective perturbations. Fig.~\ref{fig:cwOnAveKL_output} shows cwOnAveKL for output perturbation under various model parameters as functions of the noise strength $\sigma_\eta$.
For output perturbation, the cwOnAveKL decreases monotonically with $\sigma_\eta$, which is consistent with intuition. 

Fig.~\ref{fig:cwOnAveKL} shows the cwOnAveKL for objective perturbation under various model parameters as a function of the noise strength $\sigma_\eta$. In this case, the cwOnAveKL is not monotonic; increasing the noise can even lead to larger values, which is counterintuitive.
As the privacy noise increases, the estimator becomes unstable as it approaches the condition~\eqref{eq:AT}, as reflected in the increase of the generalization error  (Fig.~\ref{fig:Gen_err_Gauss_replica}). This instability can also be interpreted as high sensitivity of the estimator to small perturbations in the data \citep{obuchi2019cross}.
Therefore, although increasing the noise might be expected to enhance privacy, excessively large noise does not improve privacy in this setting. Instead, it induces instability in the estimator, limiting further improvement of privacy performance.

For both perturbation schemes, the cwOnAveKL increases as the regularization parameter $\lambda$ decreases, the sparsity level $\rho$ increases, and $\alpha$ decreases. 
This indicates that privacy is harder to maintain when the underlying signal is dense and the estimation problem becomes more difficult.

\subsection{Trade-off between prediction and privacy}

\begin{figure}
    \centering
    \includegraphics[width=3.1in]{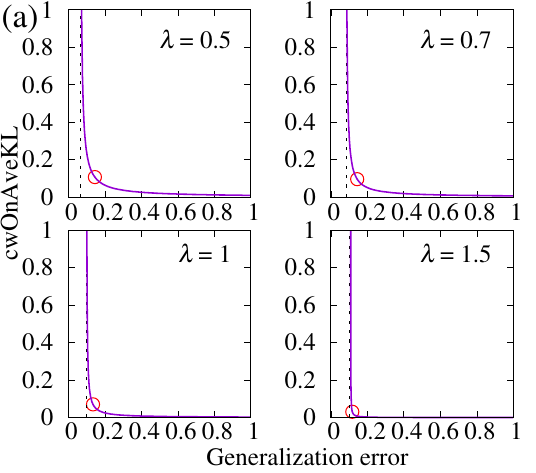}
    \includegraphics[width=3.1in]{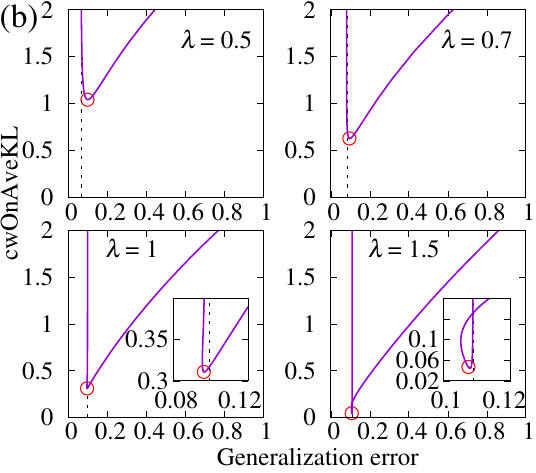}
     \caption{Privacy-accuracy trade-off at $\alpha=0.5$, $\rho=0.1$, and $\sigma_\xi=0.1$ for (a) output and (b) objective perturbation. The vertical dashed lines represent the generalization error in the noiseless case, and circles denote the points closest to the origin. The inset in (b) magnifies the curve around the minimum.}
    \label{fig:Gen_vs_OnAveKL_all}   
\end{figure}

To investigate the trade-off between prediction accuracy and privacy, we plot cwOnAveKL against the generalization error, as shown in Fig.~\ref{fig:Gen_vs_OnAveKL_all}: (a) is for output perturbation, and (b) is for objective perturbation. This figure is obtained by combining the results in Fig.~\ref{fig:vs_sigma_eta}(b) and Figs.~\ref{fig:cwOnAveKL_output}, \ref{fig:cwOnAveKL}, using the noise strength $\sigma_\eta$ as a parametric variable. Each curve represents how the prediction-privacy pair evolves as $\sigma_\eta$ varies.

In Fig.~\ref{fig:Gen_vs_OnAveKL_all}(a), the relationship between cwOnAveKL and the generalization error under output perturbation is monotonic. The limit $\sigma_\eta \to 0$ corresponds to divergence of cwOnAveKL, as the curve approaches the vertical dashed line indicating the generalization error in the absence of privacy noise. As $\sigma_\eta$ increases, the curve is traversed from this divergent regime toward the regime where cwOnAveKL vanishes. Furthermore, as $\lambda$ increases, the curve approaches a limiting shape defined by the three points $(E_0,\infty)$, $(E_0,0)$, and $(\infty,0)$.

In contrast, Fig.~\ref{fig:Gen_vs_OnAveKL_all}(b) shows that the relationship between cwOnAveKL and the generalization error in the objective perturbation case is generally non-monotonic. Starting from $\sigma_\eta = 0$, increasing $\sigma_\eta$ initially reduces cwOnAveKL while keeping the generalization error nearly unchanged or slightly decreasing it. For larger $\sigma_\eta$, both cwOnAveKL and the generalization error increase, approximately in a linear manner.
As in the output perturbation case, the minimum of the curve approaches $(E_0,0)$ as $\lambda$ increases; however, the divergence at large $\sigma_\eta$ does not disappear.

From a practical viewpoint, desirable noise strengths are those for which both the generalization error and cwOnAveKL are small. Motivated by this observation, we define the \emph{optimal privacy noise} as the value of $\sigma_\eta$ that minimizes the distance to the origin in the generalization-cwOnAveKL plane, i.e., the point that simultaneously achieves low generalization error and strong privacy. These optimal points are indicated by circles in Fig.~\ref{fig:Gen_vs_OnAveKL_all}.
For output perturbation, beyond the optimal noise level, privacy continues to improve while the generalization error deteriorates. For objective perturbation, beyond the optimal noise level, further increasing $\sigma_\eta$ causes both cwOnAveKL and the generalization error to grow approximately proportionally for any $\lambda$.

\begin{figure}
\centering
\begin{minipage}{0.495\textwidth}
        \centering
    \includegraphics[width=0.8\linewidth]{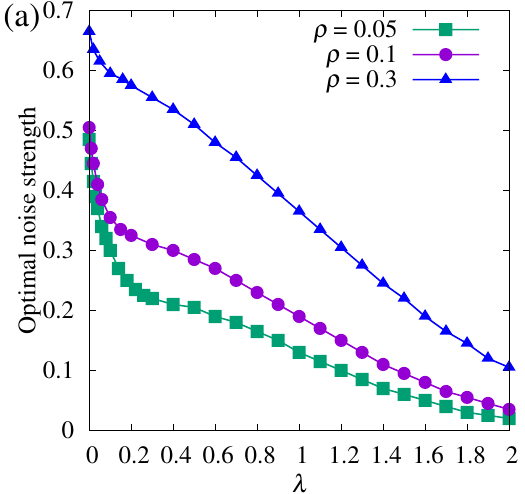}
\end{minipage}
\begin{minipage}{0.495\textwidth}
        \centering
    \includegraphics[width=0.8\linewidth]{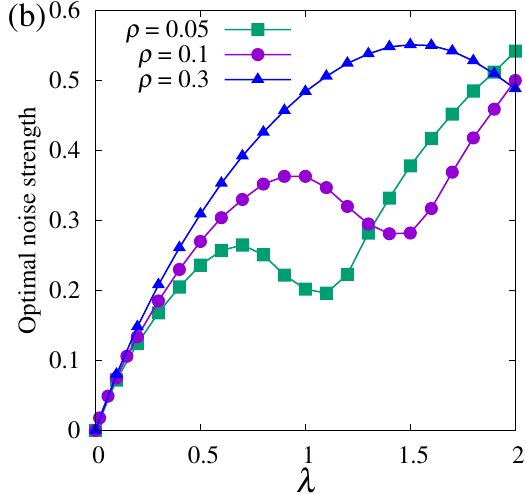}
\end{minipage}
   \caption{(a) Optimal noise strength under output perturbation at $\alpha=0.5$ and $\sigma_\xi=0.1$ for different values of $\rho$. (b) Optimal noise strength under objective perturbation under the same conditions.}
    \label{fig:optimal_noise}
\end{figure}

\begin{figure}
\centering
\begin{minipage}{0.495\textwidth}
      \centering
    \includegraphics[width=0.8\linewidth]{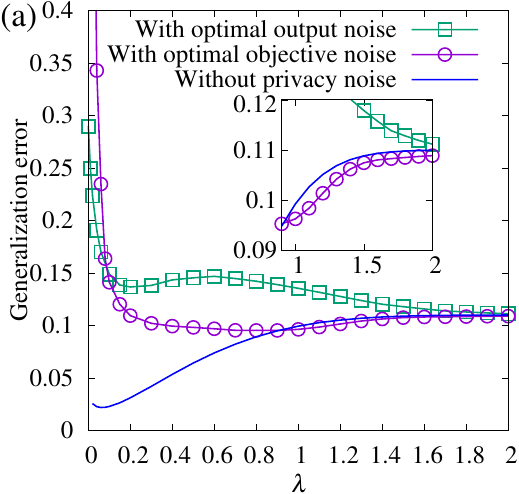}  
\end{minipage}
\begin{minipage}{0.495\textwidth}
\centering
    \includegraphics[width=0.8\linewidth]{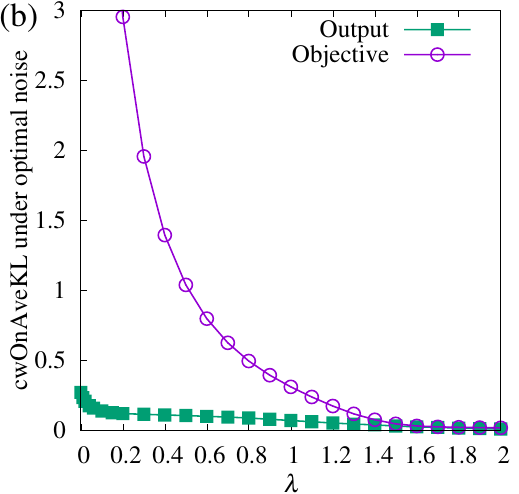}    
\end{minipage}
    \caption{Generalization error (a) and cwOnAveKL (b) at the optimal noise strength, for $\alpha=0.5$, $\rho=0.1$, and $\sigma_\xi=0.1$.}
    \label{fig:gen_and_privacy_at_optimal}
\end{figure}

Figure~\ref{fig:optimal_noise} shows the dependence of the optimal noise level on the regularization parameter $\lambda$ at $\alpha=0.5$ and $\sigma_\xi=0.1$ for (a) output and (b) objective perturbation.
For output perturbation, the optimal noise decreases as $\lambda$ increases, and stronger noise is required for larger $\rho$. In contrast, for objective perturbation, the optimal noise exhibits a non-monotonic dependence on $\lambda$.
As discussed above, beyond the optimal noise level, both the privacy and generalization performance deteriorate. Therefore, in the objective perturbation case, a more careful tuning of the noise strength is required.

Figure~\ref{fig:gen_and_privacy_at_optimal} shows (a) the generalization error and (b) cwOnAveKL evaluated at the optimal noise level.
For generalization, objective perturbation achieves lower error over a wider range of $\lambda$, and in particular outperforms the noiseless case at large $\lambda$. In contrast, cwOnAveKL is smaller for output perturbation across the entire range of $\lambda$.
These results suggest that when strong privacy is not required, objective perturbation may be preferable, as it maintains or even improves generalization performance. 
On the other hand, when a higher level of privacy is desired, output perturbation may be more suitable, although it slightly reduces prediction accuracy.
We emphasize that this conclusion relies on the assumption that the support selection remains unchanged under one-point perturbations of the data, which may be a reasonable approximation in the high-dimensional limit considered in this paper.

\section{Conclusion}

In this work, we investigated statistical estimation accuracy and privacy using the on-average KL criterion in high-dimensional LASSO with output and objective perturbation. We quantified how privacy noise affects estimation behavior. In the case of objective perturbation, the estimator can become denser than in the noiseless setting, even under the same regularization strength. This property can lead to improved generalization performance compared to the noiseless case, particularly for sufficiently large $\lambda$ and appropriately chosen noise levels.

We analyzed the distributional properties of the estimator with respect to privacy noise using asymptotic analysis based on approximate message passing (AMP) and its associated state evolution. This framework allowed us to derive a statistically equivalent characterization of the AMP estimator and to quantify how the privacy level, measured by cwOnAveKL, depends on the noise strength and model parameters.

Our analysis revealed several notable features. First, the privacy level is not a monotonic function of the objective perturbation noise. While adding output noise consistently improves privacy, excessively large objective noise can instead increase cwOnAveKL. This counterintuitive behavior arises from instability in the estimator, as captured by AMP, which is closely related to coordinate descent dynamics, and is reflected in increased generalization error. In such regimes, the estimator becomes highly sensitive to data perturbations, and the privacy protection deteriorates.

Second, cwOnAveKL depends strongly on the underlying model parameters. In particular, weaker regularization, lower sampling ratios, and denser true signal tend to increase cwOnAveKL. These findings suggest that privacy is harder to maintain precisely in regimes where statistical estimation itself is challenging, while sparsity can play a beneficial role in enhancing privacy.

To better understand the trade-off between prediction accuracy and privacy, we examined the relationship between cwOnAveKL and the generalization error by treating the privacy noise as a parametric variable. In the case of output perturbation, the resulting curve exhibits an inverse relationship, whereas for objective perturbation it becomes non-monotonic. This indicates that moderate noise levels can simultaneously improve prediction performance and provide privacy protection, while excessive noise degrades both. Based on this observation, we introduced the notion of an \emph{optimal privacy noise}, defined as the noise strength that minimizes the distance to the origin in the generalization-cwOnAveKL plane. This point represents a balance between prediction accuracy and privacy. We further found that the optimal noise decreases monotonically with $\lambda$ for output perturbation, while it exhibits non-monotonic behavior for objective perturbation. Moreover, objective perturbation tends to better preserve prediction performance, whereas output perturbation provides stronger privacy protection in terms of cwOnAveKL.

Overall, our results highlight that privacy protection through noise injection must be carefully calibrated. Increasing objective noise does not necessarily improve privacy and may instead induce estimator instability and degrade prediction performance. The proposed framework provides a quantitative approach to analyzing this trade-off and offers guidance for selecting noise levels in high-dimensional statistical models.

Future work includes extending this analysis to broader classes of models and inference methods, as well as to other privacy mechanisms beyond output and objective perturbation. In particular, it would be important to investigate data-dependent noise calibration strategies and to validate these findings in more complex statistical settings. Connections to differential privacy mechanisms such as output perturbation, objective perturbation, and gradient-based methods \citep{abadi2016deep} provide a promising direction for further study.

\bigskip
\begin{center}
{\large\bf SUPPLEMENTARY MATERIAL}
\end{center}

Supplementary material contains proof and derivations of theorem and claims.

\appendix
\renewcommand{\thesection}{S.\arabic{section}}
\setcounter{section}{0}

\section{AMP algorithm}

\begin{figure}
    \centering
    \includegraphics[width=4in]{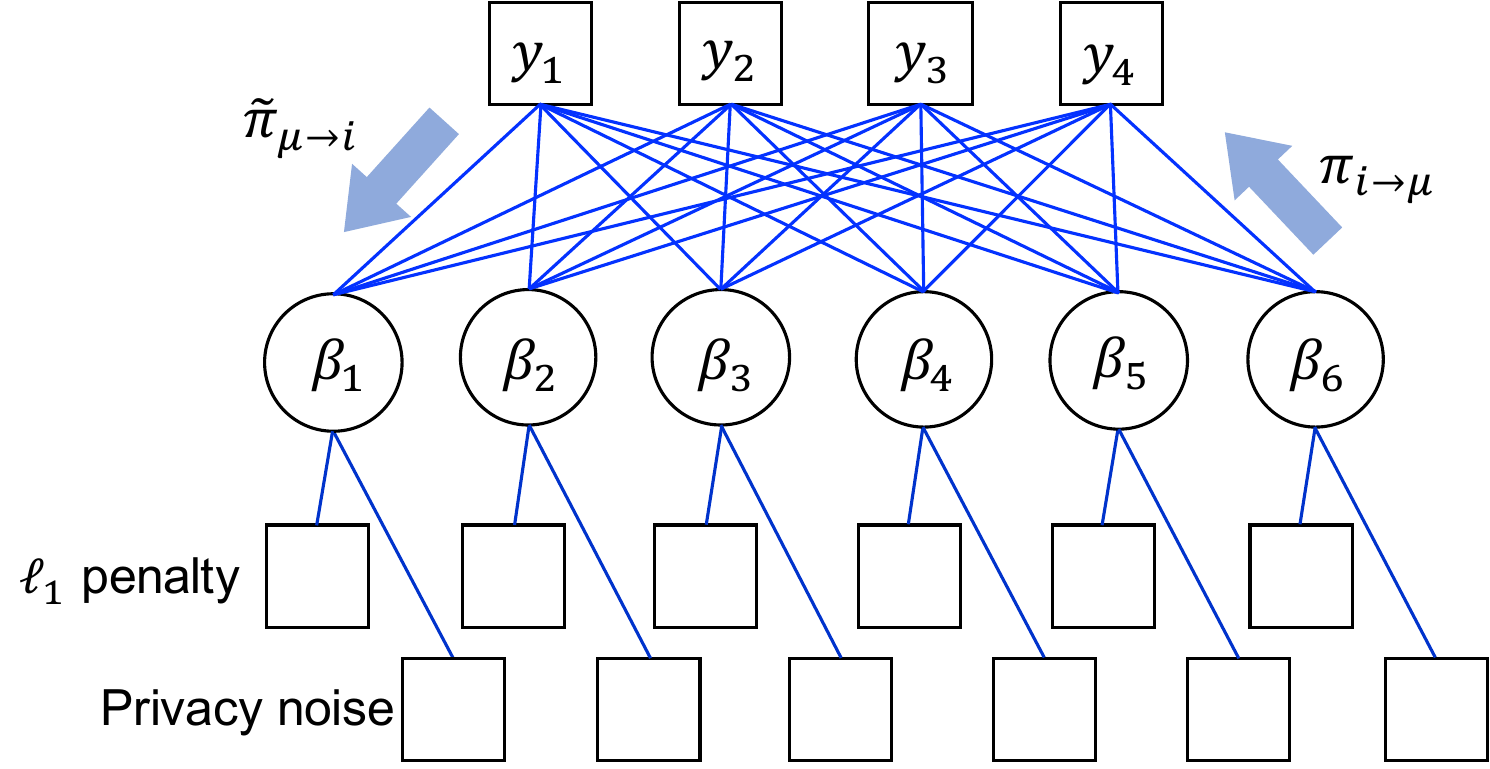}
    \caption{Graphical representation of the privacy-preserving LASSO with $n=4$ and $p=6$.}
    \label{fig:factor_graph_PrivacyLASSO}
\end{figure}

Fig.~\ref{fig:factor_graph_PrivacyLASSO} shows the factor graph representation of the posterior distribution with privacy noise, given by Eq.~(10) in the main text. In this representation:
\begin{itemize}
    \item Variable nodes correspond to the parameters $\bm{\beta}$.
    \item Factor nodes correspond to likelihood terms associated with each observation $y_\mu$; the $\ell_1$ penalty and the privacy noise are also represented as factor nodes.
    \item Edges connect factor and variable nodes whenever the corresponding factor depends on the variable.
\end{itemize}
Messages are assigned to edges as follows:
\begin{itemize}
    \item Output message $\pi_{i\to\mu}(\beta_i)$: Message from the variable $\beta_i$ to data $y_\mu$
    \item Input message $\widetilde{\pi}_{\mu\to i}(\beta_i)$: Message from data $y_\mu$ to the variable $\beta_i$.
\end{itemize}
By interpreting the privacy noise as an additional prior on the variables, the standard procedure for deriving AMP can be applied, following \citep{sakata2023prediction}. The resulting marginal posterior is given by Eq.~(21) in the main text.

Introducing appropriate mean and variance parameters, we can express both the marginal posterior and the messages by using eq.(21) in the main text. In particular, we introduce the following representations, which are useful for the analysis of the subsequent state evolution:
\begin{align}
{\Sigma}^{(t)}_{i\to\mu}=\frac{1}{\sum_{\nu\in\bm{M}\backslash \mu}
(\widetilde{s}^{(t)}_{\nu\to i})^{-1}},~~~
\mathrm{m}^{(t)}_{i\to\mu}=\frac
{\sum_{\nu\in\bm{M}\backslash \mu}(\widetilde{s}^{(t)}_{\nu\to i})^{-1}
\widetilde{\beta}^{(t)}_{\nu\to i}}{\sum_{\nu\in\bm{M}\backslash \mu}(\widetilde{s}^{(t)}_{\nu\to i})^{-1}}.\label{eq:output_macro_param}
\end{align}
and 
\begin{align}
{\Sigma}^{(t)}_{i}=\frac{1}{\sum_{\nu=1}^M(\widetilde{s}^{(t)}_{\nu \to i})^{-1}},~~~
\mathrm{m}^{(t)}_{i}=\frac
{\sum_{\nu=1}^M(\widetilde{s}^{(t)}_{\nu\to i})^{-1}\widetilde{\beta}^{(t)}_{\nu \to i}}{\sum_{\nu=1}^M(\widetilde{s}^{(t)}_{\nu\to i})^{-1}}.
\end{align}
Under the Gaussian approximation, the input message takes the form
\begin{align}
\widetilde{\xi}^{(t)}_{\mu\to i}(x_{i})&\propto 
\exp\left\{-\frac{\tau A^{(t)}_{\mu\to i}\beta_{i}^2}{2}+\tau B^{(t)}_{\mu\to i}\beta_{i}\right\},
\label{eq:input_final}
\end{align}
with 
\begin{align}
A_{\mu\to i}^{(t)}&=\frac{X^2_{\mu i}}{1+s_\theta^{(t)}}\label{eq:bp_A}\\
B_{\mu\to i}^{(t)}&=X_{\mu i}\frac{y_\mu-\widehat{y}_\mu^{\setminus\mu(t)}}{1+s_\theta^{(t)}},\label{eq:bp_B}
\end{align}
hence the mean and (rescaled) variance of the input message, $\widetilde{\beta}_{\mu\to i}$ and $\widetilde{s}_{\mu\to i}$, are given by
\begin{align}
\widetilde{\beta}^{(t)}_{\mu\to i}&=\frac{B^{(t)}_{\mu\to i}}{A^{(t)}_{\mu\to i}},\\
\widetilde{s}^{(t)}_{\mu\to i}&=\frac{1}{A^{(t)}_{\mu\to i}}.\label{eq:input_stat}
\end{align}

\section{State Evolution}
\label{sec:app_DE}

We start from the following expressions, obtained from Eqs.~\eqref{eq:bp_A}--\eqref{eq:input_stat}:
\begin{align}
\mathrm{m}_{i}^{(t)}&={\Sigma}_i^{(t)}\sum_{\mu=1}^n\left\{
X_{\mu i}\frac{y_\mu-\widehat{y}_\mu^{\setminus}}{1+{s_\theta}}
+\frac{X_{\mu i}^{2}\widehat{\beta}_{i}^{(t)}}{1+{s_\theta}}\right\}\\
\frac{1}{{\Sigma}_i^{(t)}}&=
\sum_{\mu=1}^n\frac{X_{\mu i}^2}{1+{s_\theta}}.
\end{align}
In the large-system limit, the fluctuations of ${\Sigma}_{i}^{(t)}$ vanish, and ${\Sigma}_{i}^{(t)}$ converges to a deterministic value.
Consequently, the only remaining random quantity that requires a distributional
characterization is $\mathrm{m}_i^{(t)}$.
Using the definition of the leave-one-out prediction
$\widehat{y}_\mu^{\setminus\mu}$ and the data model
$y_\mu=\sum_{j=1}^pX_{\mu j}\beta_j^{(0)}+\xi_\mu$,
we obtain
\begin{align}
\nonumber
\mathrm{m}_{i}^{(t)}
\nonumber
&=\frac{1}{\sum_{\mu=1}^nX^{2}_{\mu i}\displaystyle\frac{1}{1+s_\theta}}\sum_{\mu=1}^n\left\{
X_{\mu i}\displaystyle\frac{(y_\mu-\widehat{y}_\mu^{\setminus\mu})}{1+s_\theta}
+X_{\mu i}^{2}\widehat{\beta}_{i\to\mu}^{(t)}
\frac{1}{1+s_\theta}\right\}\\
\nonumber
&=\frac{1}{\sum_{\mu=1}^nX^{2}_{\mu i}}\sum_{\mu=1}^n\left\{
X_{\mu i}(\sum_{i=1}^pX_{\mu i}\beta_i^{(0)}+\xi_\mu-\sum_{i=1}^pX_{\mu i}\widehat{\beta}_{i\to\mu}^{(t)})
+X_{\mu i}^{2}\widehat{\beta}_{i\to\mu}^{(t)}
\right\}\\
&=\frac{1}{\sum_{\mu=1}^nX^{2}_{\mu i}}\sum_{\mu=1}^n\left\{
X_{\mu i}(\sum_{i=1}^pX_{\mu i}\beta_i^{(0)}+\xi_\mu)-\sum_{j\neq i}^pX_{\mu i}X_{\mu j}\widehat{\beta}_{j\to\mu}^{(t)}
\right\}
\end{align}
Taking the expectation with respect to $X$ and the observation noise 
$\bm{\xi}$ under the setting $\textbf{D}$, we find
\begin{align}
    \mathbb{E}_{X,\bm{\xi}}[\mathrm{m}_i]=\beta_i^{(0)}.
\end{align}

Next, we evaluate the variance of $\mathrm{m}_i^{(t)}$.
Subtracting $\beta_i^{(0)}$ and rearranging terms yields
\begin{align}
\nonumber
\mathrm{m}_{i}^{(t)}-\beta_i^{(0)}&=\frac{1}{\sum_{\mu=1}^nX^{2}_{\mu i}}
\sum_{\mu=1}^n\left\{
X_{\mu i}(\sum_{j=1}^pX_{\mu j}\beta_j^{(0)}+\xi_\mu)-\sum_{j\neq i}^pX_{\mu i}X_{\mu j}\widehat{\beta}_{j\to\mu}^{(t)}
-X_{\mu i}^2\beta_i^{(0)}\right\}\\
&=\frac{1}{\sum_{\mu=1}^nX^{2}_{\mu i}}\sum_{\mu=1}^n\left\{
X_{\mu i}(\sum_{j\neq i}^pX_{\mu j}\beta_j^{(0)}+\xi_\mu)-\sum_{j\neq i}^pX_{\mu i}X_{\mu j}\widehat{\beta}_{j\to\mu}^{(t)}
\right\}.
\end{align}
Assuming independence of $X_{\mu i}$ across indices and using the law of large
numbers, we obtain
\begin{align}
\nonumber
&\mathbb{E}_X[(\mathrm{m}_{i}^{(t)}-\beta_i^{(0)})^2]
\simeq\frac{1}{\alpha^2}\sum_{\mu=1}^n\left\{
X_{\mu i}(\sum_{j\neq i}^pX_{\mu j}\beta_j^{(0)}+\xi_\mu)-\sum_{j\neq i}^pX_{\mu i}X_{\mu j}\widehat{\beta}_{j\to\mu}^{(t)}
\right\}^2\\
\nonumber
&\simeq\frac{1}{\alpha}\left\{
(\sum_{j\neq i}^pX_{\mu j}^2(\beta_j^{(0)})^2+\xi_\mu^2)-2\sum_{k\neq i}X_{\mu k}^2\beta_k^{(0)}\widehat{\beta}_{k\to\mu}^{(t)}+\sum_{j\neq i}^pX_{\mu j}^2(\widehat{\beta}_{j\to\mu}^{(t)})^2
\right\}\\
\nonumber
&\simeq\frac{1}{\alpha}\left\{
(\frac{1}{p}\sum_{j\neq i}^p(\beta_j^{(0)})^2+\sigma_\xi^2)-\frac{2}{p}\sum_{k\neq i}\beta_k^{(0)}\widehat{\beta}_{k\to\mu}^{(t)}+\frac{1}{p}\sum_{j\neq i}^p(\widehat{\beta}_{j\to\mu}^{(t)})^2
\right\}\\
&\simeq\frac{1}{\alpha}\left\{\frac{1}{p}\sum_{j\neq i}\left(\beta_j^{(0)}-\widehat{\beta}_j^{(t)}\right)^2+\sigma_\xi^2\right\}=\frac{E^{(t)}}{\alpha},
\label{eq:app_m_variance}
\end{align}
where $E^{(t)}$ denotes the state evolution parameter corresponding to the generalization error at iteration $t$.

Therefore, in the large-system limit, $\mathrm{m}_i^{(t)}$ converges in distribution to a Gaussian random variable:
\begin{align}
    \mathrm{m}_i^{(t)}
    \;\xrightarrow{d}\;
    {\cal N}\!\left(\beta_i^{(0)},\,\sigma_z^2\right),
    \qquad
    \sigma_z^2=\frac{E^{(t)}}{\alpha}.
\end{align}
This Gaussianity of the effective field $\mathrm{m}_i^{(t)}$ leads directly to
the state evolution equation and establishes the decoupling principle used in
Theorem~4.1 of the main text.

\subsection{Distribution of estimator with respect to noise}
\label{sec:app_beta_dist}

To derive the distribution of the estimator $\beta$ under privacy noise, 
we separately consider the cases $\beta>0$, $\beta<0$, and $\beta=0$. 
Here, we define $\widehat{\mathrm{m}}_i = \beta_i^{(0)} + \sigma_z z$.
For $\beta>0$, i.e., when 
$\widehat{\mathrm{m}}_i - \eta \Sigma > \lambda \Sigma$, 
the conditional distribution is given by
\begin{align}
\nonumber
    \mathbb{P}^{\mathrm{SE}}_{|\beta^{(0)},z}(\beta|\beta>0;\widehat{\mathrm{m}}_i,\Sigma)d\beta&=\int d\eta P_\eta(\eta)\mathbb{I}\left(\beta\leq\widehat{\mathrm{m}}_i-\eta_i\Sigma-\lambda\Sigma
<\beta+d\beta\right)\\
&=\frac{1}{2}\left\{\mathrm{erfc}\left(-\frac{\beta+d\beta-\widehat{\mathrm{m}}_i+\lambda\Sigma}{\sqrt{2}\Sigma\sigma_\eta}\right)-\mathrm{erfc}\left(-\frac{\beta-\widehat{\mathrm{m}}_i+\lambda\Sigma}{\sqrt{2}\Sigma\sigma_\eta}\right)\right\}
\end{align}
For sufficiently small $d\beta$, this yields
\begin{align}
\mathbb{P}^{\mathrm{SE}}_{|\beta^{(0)},z}(\beta|\beta>0;\widehat{\mathrm{m}}_i,\Sigma)d\beta
&=\frac{1}{\sqrt{2\pi}\Sigma\sigma_\eta}\exp\left(-\frac{(\beta-\widehat{\mathrm{m}}_i+\lambda\Sigma)^2}{2\Sigma^2\sigma_\eta^2}\right)d\beta+O(d\beta^2).
\end{align}
Similarly, for $\beta<0$, i.e., when 
$\widehat{\mathrm{m}}_i - \eta \Sigma < -\lambda \Sigma$, 
we obtain
\begin{align}
\nonumber
\mathbb{P}^{\mathrm{SE}}_{|\beta^{(0)},z}(\beta|\beta<0;\widehat{\mathrm{m}}_i,\Sigma)d\beta
&=\frac{1}{\sqrt{2\pi}\Sigma\sigma_\eta}\exp\left(-\frac{(\beta-\widehat{\mathrm{m}}_i-\lambda\Sigma)^2}{2\Sigma^2\sigma_\eta^2}\right)d\beta+O(d\beta^2).
\end{align}
The probability that the estimator is nonzero is therefore given by
\begin{align}
\nonumber
    \widehat{r}^{\mathrm{SE}}(\widehat{\mathrm{m}}_i,\Sigma)&=\int_{\beta=0}^\infty \mathbb{P}^{\mathrm{SE}}_{|\beta^{(0)},z}(\beta|\beta>0;\widehat{\mathrm{m}}_i,\Sigma)d\beta+\int_{-\infty}^0\mathbb{P}^{\mathrm{SE}}_{|\beta^{(0)},z}(\beta|\beta<0;\widehat{\mathrm{m}}_i,\Sigma)d\beta\\
    &=\frac{1}{2}\left\{\mathrm{erfc}\left(\frac{-\widehat{\mathrm{m}}_i+\lambda\Sigma}{\sqrt{2}\Sigma\sigma_\eta}\right)+\mathrm{erfc}\left(\frac{\widehat{\mathrm{m}}_i+\lambda\Sigma}{\sqrt{2}\Sigma\sigma_\eta}\right)\right\},
\end{align}
In summary, this yields the distribution given in Eq.~(52) of the main text.

\section{On-average KL Privacy using AMP}

\subsection{cwOnAveKL for output perturbation}

To compute cwOnAveKL for output perturbation (eq.~(59) in the main text),
we evaluate the squared difference between the estimators under one-point-mutant datasets using AMP.
Using the AMP expression, we obtain
\begin{align}
\nonumber
&\sum_{i=1}^p(\widehat{\beta}_{\mathrm{AMP},i}(\{{\cal D}_{\setminus\mu},\bm{d}_\mu\},\bm{\eta})-\widehat{\beta}_{\mathrm{AMP},i}(\{{\cal D}_{\setminus\mu},\bm{d}^\prime_\mu\},\bm{\eta}))^2\\
\nonumber
&=\sum_{i=1}^p(\widehat{\beta}_{\mathrm{AMP},i}(\{{\cal D}_{\setminus\mu},\bm{d}_\mu\},\bm{\eta})-\widehat{\beta}_{\mathrm{AMP},i}({\cal D}_{\setminus\mu},\bm{\eta})-(\widehat{\beta}_{\mathrm{AMP},i}(\{{\cal D}_{\setminus\mu},\bm{d}_\mu^\prime\},\bm{\eta})-\widehat{\beta}_{\mathrm{AMP},i}({\cal D}_{\setminus\mu},\bm{\eta})))^2\\
&=\sum_{i=1}^p\left({s}_{i}X_{\mu i}\frac{(y_\mu-\widehat{y}_\mu^{\setminus\mu})}{1+s_\theta}-{s}_{i}X_{\mu i}^\prime\frac{(y_\mu^\prime-\widehat{y}_\mu^{\prime\setminus\mu})}{1+s_\theta}\right)^2,
\end{align}
where $\widehat{y}_\mu^{\prime\setminus\mu}=\sum_{i=1}^pX_{\mu i}^\prime\widehat{\beta}_{i\to\mu}$.
Note that $s_i=0$ for inactive components, hence the indicator function
in eq.~(59) is implicitly incorporated.

Since $X_{\mu i}$ and $X_{\mu i}^\prime$ are independent for any $i$,
taking the expectation over the dataset yields
\begin{align}
    \mathbb{E}_{\cal D}\left[\sum_{i=1}^p(\widehat{\beta}_{\mathrm{AMP},i}(\{{\cal D}_{\setminus\mu},\bm{d}_\mu\},\bm{\eta})-\widehat{\beta}_{\mathrm{AMP},i}(\{{\cal D}_{\setminus\mu},\bm{d}^\prime_\mu\},\bm{\eta}))^2\right]=\frac{2}{p}\frac{E}{(1+V)^2}\sum_{i=1}^p\left(s_i\right)^2.
\end{align}
Using eqs. (29) and (31) in the main text, 
\begin{align}
    \mathbb{E}_{\cal D}\left[\sum_{i=1}^p(\widehat{\beta}_{\mathrm{AMP},i}(\{{\cal D}_{\setminus\mu},\bm{d}_\mu\},\bm{\eta})-\widehat{\beta}_{\mathrm{AMP},i}(\{{\cal D}_{\setminus\mu},\bm{d}^\prime_\mu\},\bm{\eta}))^2\right]=\frac{2E\widehat{\rho}}{\alpha^2}
    \label{app_eq:sensitivity}
\end{align}
This corresponds to the average {\it sensitivity} of the AMP estimator under
setting \textbf{D}, which at $\bm{\eta}=\bm{0}$ leads to eq.~(60) in the main text.

\subsection{Correlation structure under one-point-mutant data}

To derive cwOnAveKL, we need to characterize the correlation between the estimators obtained from the original dataset ${\cal D}$ and the one-point-mutant dataset ${\cal D}^\prime$.
To this end, we analyze the joint distribution of the AMP mean parameters $\widehat{\mathrm{m}}_i$ and $\widehat{\mathrm{m}}_{i}^{\prime\mu}$, where $\widehat{\mathrm{m}}_{i}^{\prime\mu}$ denotes the mean parameter under the $\mu$-th one-point-mutant dataset. This analysis allows us to derive Claim~6.1.

Focusing on the overlap between $\widehat{\mathrm{m}}_i$ and $\widehat{\mathrm{m}}_{i}^{\prime\mu}$, we can rewrite them as
\begin{align}
\mathrm{m}_i&=\frac{1}{\alpha}\left\{\sum_{\nu\neq\mu}X_{\nu i}\left(y_\nu-\widehat{y}_\nu^{\setminus\nu}+X_{\nu i}\widehat{\beta}_{i\to\nu}\right)+X_{\mu i}\left(y_\mu-\widehat{y}_\mu^{\setminus\mu}+X_{\mu i}\widehat{\beta}_{i\to\mu}\right)\right\}\\
\mathrm{m}^{\prime\mu}_{i}&=\frac{1}{\alpha}\left\{\sum_{\nu\neq \mu}X_{\nu i}\left(y_\nu-\widehat{y}_\nu^{\setminus\nu}+X_{\nu i}\widehat{\beta}_{i\to\nu}\right)+X_{\mu i}^\prime\left(y_\mu^\prime-\widehat{y}_\mu^{\prime\setminus\mu}+X_{\mu i}^\prime\widehat{\beta}_{i\to\mu}\right)\right\}.
\end{align}
From the independence between $X_{\mu i}$ and $X_{\mu i}^\prime$, we obtain 
\begin{align}
    \mathbb{E}[(\mathrm{m}_i-\beta_i^{(0)})(\mathrm{m}_{i}^\prime-\beta_i^{(0)})]&=\frac{E}{\alpha}-\frac{E}{\alpha n}.
\end{align}
From Eq.~\eqref{eq:app_m_variance}, we have
$\mathbb{E}[(\mathrm{m}_i-\beta_i^{(0)})^2]=\mathbb{E}[(\mathrm{m}^{\prime\mu}_i-\beta_i^{(0)})^2]=E\slash \alpha$, hence
the joint distributional property of $\mathrm{m}_i$ and $\mathrm{m}_i^{\prime\mu}$ can be constructed as
\begin{align}
    \mathrm{m}_i-\beta_i^{(0)}&=\sigma_z^{\setminus\Delta}z+\sqrt{\Delta}\zeta\\
    \mathrm{m}_i^\prime-\beta_i^{(0)}&=\sigma_z^{\setminus\Delta}z+\sqrt{\Delta}\zeta^\prime
\end{align}
where $z$, $\zeta$, and $\zeta^\prime$ are independent standard Gaussian random variables.
These expressions imply the distributional equivalence
\begin{align}
    f(\widehat{\mathrm{m}}_i^{\prime\mu},\widehat{\mathrm{m}}_i)\overset{d}{=}f(\beta_i^{(0)}+\sigma_z^{\setminus\Delta}z+\sqrt{\Delta}\zeta^\prime,\beta_i^{(0)}+\sigma_z^{\setminus\Delta}z+\sqrt{\Delta}\zeta),
\end{align}
for any $\mu$.
By choosing $f$ as a functional of the corresponding marginal distributions $a(\mathbb{P}(\widehat{\mathrm{m}}),\mathbb{P}(\widehat{\mathrm{m}}^\prime))$, this establishes Claim~6.1.

\subsection{Component-wise on-average KL}
\label{sec:app_cwOnAveKL}

We evaluate the component-wise on-average KL divergence between
SE-based distributions.
The cross entropy between
$\mathbb{P}_{|\beta^{(0)},z}^{(\mathrm{SE})}(\beta|\widehat{\mathrm{m}}_1,\Sigma_1)$
and
$\mathbb{P}^{(\mathrm{SE})}_{|\beta^{(0)},z}(\beta|\widehat{\mathrm{m}}_2,\Sigma_2)$
is given by
\begin{align}
\nonumber
&\int d\beta \mathbb{P}_{|\beta^{(0)},z}^{(\mathrm{SE})}(\beta|\widehat{\mathrm{m}}_1,\Sigma_1)\ln\mathbb{P}_{|\beta^{(0)},z}^{(\mathrm{SE})}(\beta|\widehat{\mathrm{m}}_2,\Sigma_2)\\
\nonumber
&=\left(1-\widehat{r}(\widehat{\mathrm{m}}_1,\Sigma_1)\right)\ln\left(1-\widehat{r}(\widehat{\mathrm{m}}_2,\Sigma_2)\right)-\frac{\widehat{r}(\widehat{\mathrm{m}}_1,\Sigma_1)}{2}\ln\left(2\pi\Sigma_2^2\sigma_\eta^2\right)\\
&+\frac{1}{\sqrt{2\pi\Sigma_1^2\sigma_\eta^2}}\int d\beta\left\{-\frac{(\beta-\widehat{\mathrm{m}}_2+\mathrm{sgn}(\beta)\lambda\Sigma_2)^2}{2\Sigma_2^2\sigma_\eta^2}\right\}\exp\left\{-\frac{(\beta-\widehat{\mathrm{m}}_1+\mathrm{sgn}(\beta)\lambda\Sigma_1)^2}{2\Sigma_1^2\sigma_\eta^2}\right\}.
\label{eq:app_cross_entropy}
\end{align}
We now consider the case where
$\widehat{\mathrm{m}}_1=\widehat{\mathrm{m}}^{\setminus\Delta}+\sqrt{\Delta}\zeta$
and
$\widehat{\mathrm{m}}_2=\widehat{\mathrm{m}}^{\setminus\Delta}+\sqrt{\Delta}\zeta'$,
with
$\widehat{\mathrm{m}}^{\setminus\Delta}=\beta^{(0)}+\sigma_z^{\setminus\Delta}z$,
and $\Sigma_1=\Sigma_2=\Sigma$.
Writing
$\widehat{\mathrm{m}}_1=\widehat{\mathrm{m}}_2+\sqrt{\Delta}(\zeta-\zeta')$,
we expand the cross entropy in powers of $\sqrt{\Delta}$ up to
$O(\Delta)$.
In this derivation, we use the following identities:
\begin{align}
     \widehat{r}(\mathrm{m},\Sigma)&=\frac{1}{\sqrt{2\pi\Sigma^2\sigma_\eta^2}}\int d\beta\exp\left\{-\frac{(\beta-\mathrm{m}+\mathrm{sgn}(\beta)\lambda\Sigma)^2}{2\Sigma^2\sigma_\eta^2}\right\}\\
    \frac{\partial\widehat{r}(\mathrm{m},\Sigma)}{\partial\mathrm{m}}&=\frac{1}{\sqrt{2\pi\Sigma^2\sigma_\eta^2}}\int d\beta\left(\frac{\beta-\mathrm{m}+\mathrm{sgn}(\beta)\lambda\Sigma}{\Sigma^2\sigma_\eta^2}\right)\exp\left(-\frac{(\beta-\mathrm{m}+\mathrm{sgn}(\beta)\lambda\Sigma)^2}{2\Sigma^2\sigma_\eta^2}\right)\\
   \frac{\partial^2\widehat{r}(\mathrm{m},\Sigma)}{\partial\mathrm{m}^2}&=-\frac{\widehat{r}}{\Sigma^2\sigma_\eta^2}+\frac{1}{\sqrt{2\pi\Sigma^2\sigma_\eta^2}}\int d\beta\left(\frac{\beta-\mathrm{m}+\mathrm{sgn}(\beta)\lambda\Sigma}{\Sigma^2\sigma_\eta^2}\right)^2\exp\left(-\frac{(\beta-\mathrm{m}+\mathrm{sgn}(\beta)\lambda\Sigma)^2}{2\Sigma^2\sigma_\eta^2}\right).
\end{align}
Combining the above expressions, we arrive at Eq.~(63) in the main text.

\section{Replica method}
\label{sec:app_replica}

To analytically characterize the typical performance of the privacy LASSO estimator in the large-system limit, we employ the replica method, a non-rigorous yet powerful analytical technique from statistical physics that enables the evaluation of ensemble-averaged quantities in high-dimensional inference problems. This framework yields closed-form expressions for key performance metrics, including the training and generalization errors.

\subsection{Self-averaging of the noise-dependent free energy}

The property in Eq.~(11) of the main text naturally aligns with the structure of the replica method.
The central object of interest is the \emph{free energy density}, which summarizes the macroscopic behavior of the posterior distribution defined in Eq.~(10) of the main text.
For a fixed realization of the privacy noise $\bm{\eta}$ and finite $p$, we define the free energy density as
\begin{align}
    f(\bm{\eta}) = -\lim_{\tau\to\infty}\frac{1}{p\tau}\mathbb{E}_{\cal D}[\ln Z_\tau({\cal D},\bm{\eta})].
\end{align}
To evaluate the large dimensional limit of $f_p(\bm{\eta})$, we rely on the replica
method, which is based on the identity
\begin{align}
\mathbb{E}_{\mathcal D}[\ln Z_\tau(\mathcal D,\bm{\eta})]
= 
\lim_{r\to 0}
\frac{\partial}{\partial r}
\mathbb{E}_{\mathcal D}\!\left[
    Z_\tau^r(\mathcal D,\bm{\eta})
\right].
\label{eq:lnZ_limit}
\end{align}
Assuming that $r$ is a positive integer, 
we represent $Z_\tau^r$ as an
$r$-replicated system by introducing $r$ copies of the original variable
$\bm{\beta}$, denoted by $\bm{\beta}^{(1)},\cdots,\bm{\beta}^{(r)}$,
\begin{align}
\mathbb{E}_{\cal D}[Z_\tau^r({\cal D},\bm{\eta})]=\mathbb{E}_{\cal D}\left[
\int d\bm{\beta}^{(1)}\cdots d\bm{\beta}^{(r)}
\exp\left\{-\tau\sum_{a=1}^r\left(\frac{1}{2} \|\bm{y}-X\bm{\beta}^{(a)}\|_2^2+\lambda\|\bm{\beta}^{(a)}\|_1+\bm{\eta}^\top\bm{\beta}^{(a)}\right)\right\}\right].
\label{eq:Z_finite_n}
\end{align}
Introducing the following statistics,
which are overlap quantities commonly referred to as replica order parameters,
\begin{align}
    q^{(ab)}&=\frac{1}{p}\sum_{i=1}^p\beta_i^{(a)}\beta_i^{(b)},    \qquad
    a,b\in\{1,\ldots,r\},\label{eq:def_q_ab}\\
    m^{(a)}&=\frac{1}{p}\sum_{i=1}^p\beta_i^{(0)}\beta_i^{(a)},\qquad a\in\{1,\ldots,r\} \label{eq:def_m_a}
\end{align}
the replicated partition function can be evaluated by taking expectations with
respect to the predictor matrix $X$.
After standard manipulations (see, e.g., [Krzakala et al., 2012, Sakata, 2023]), the leading contribution to
$\mathbb{E}_{\cal D}[Z_\tau^r({\cal D},\bm{\eta})]$
in the high-dimensional limit is expressed as an extremum over the macroscopic variables
${\cal Q}:=\{q^{(ab)}\}$ and $\bm{m}:=\{m^{(a)}\}$:
\begin{align}
    \mathbb{E}_{\cal D}[Z_\tau^r({\cal D},\bm{\eta})]\propto\exp\left\{p~\mathop{\mathrm{extr}}_{{\cal Q},\bm{m}}\Phi_r(\{q^{(ab)}\},\{m^{(a)}\};\bm{\eta})\right\}.
\end{align}
Here, $\Phi_r$ denotes the free entropy density, and
$\mathop{\mathrm{extr}}_{{\cal Q},\bm{m}}$ indicates extremization with respect to
${\cal Q}$ and $\bm{m}$. The explicit form of $\Phi_r$ in the absence of privacy noise is given in [Krzakala et al., 2012, Sakata, 2023], and the present case differs by the additional noise term through the prior distribution.

To proceed, we adopt the replica-symmetric (RS) assumption, under which the extremum
is invariant under permutations of the replica indices. This assumption
enables the analytic continuation of $\Phi_r$ from integer $r$ to real $r$, followed by the limit $r\to 0$. Under the RS ansatz, we set
$q^{(ab)}=q$ for all $a\neq b$, $q^{(aa)}=Q$, and $m^{(a)}=m$ for all
$a\in\{1,\ldots,r\}$.

Denoting the resulting RS free energy by $f_p^{(\mathrm{RS})}(\bm{\eta})$, we obtain the following property.
\begin{claim}[Self-averaging of the privacy-noise-dependent free energy]
Under Assumptions~\textbf{D} and~\textbf{P}, and assuming Lipschitz continuity of the free energy
with respect to the privacy noise,
\begin{align}
    f_p^{(\mathrm{RS})}(\bm{\eta})
    \xrightarrow{p}
    \mathbb{E}_\eta[f_p^{(\mathrm{RS})}(\bm{\eta})].
    \label{eq:f_self_averaging}
\end{align}
\end{claim}
\begin{Rem}
Although the present analysis is carried out under the RS assumption, the self-averaging property of the free energy is expected to hold more generally, including under one-step and multi-step replica-symmetry-breaking (RSB)
assumptions, which are commonly regarded as providing progressively more
accurate descriptions of the underlying free energy.
\end{Rem}
Equation \eqref{eq:f_self_averaging} indicates that, while the free energy is defined for each realization of the privacy noise $\bm{\eta}$, its large-system limit becomes deterministic and depends only on the distribution $P_\eta$. Consequently, macroscopic observables derived from the free energy do not depend on a specific realization of $\bm{\eta}$.

We now present the explicit form of the RS free energy in the limit $\tau\to\infty$. 
To take this limit, we introduce the scaling $\chi=\tau(Q-q)$ and keep $\chi$ finite as $\tau\to\infty$. 
The RS free energy density is then obtained as the extremum with respect to the macroscopic variables
$\Omega=\{Q,\chi,m\}$:
\begin{align}
\nonumber
f^{(\mathrm{RS})}&=\mathop{\mathrm{extr}}_{\Omega,\widehat{\Omega}}\Big[\widehat{\mu}m-\frac{\widehat{\Theta}Q-\widehat{\chi}\chi}{2}-\frac{\alpha(Q-2m+\rho\sigma_x^2+\sigma_\xi^2)}{2(1+\chi)}\\
&\hspace{2.0cm}+\int d\eta P_\eta(\eta)\int d\beta^{(0)}\phi^{(0)}(\beta^{(0)})\int Dz {\cal H}^{(\mathrm{RS})*}(z,\beta^{(0)},\eta)\Big].
\label{eq:RS_free_energy}
\end{align}
Here, $\widehat{\Omega}=\{\widehat{\mu},\widehat{\Theta},\widehat{\chi}\}$
are the conjugate variables associated with $\Omega$, whose extremization yields
\begin{align}
\widehat{\Theta}&=\frac{\alpha}{1+\chi }\label{eq:Q_hat}\\
\widehat{\chi}&=\frac{\alpha(Q-2m+\rho\sigma_x^2+\sigma_\xi^2)}{(1+\chi)^2}\label{eq:chi_hat}\\
\widehat{\mu}&=\frac{\alpha}{1+\chi }.\label{eq:m_hat}
\end{align}
We define $\beta_{\mathrm{RS}}^*$ as the minimizer of the effective single-body problem
\begin{align}
\beta_{\mathrm{RS}}^*(z,\beta^{(0)},\eta)&=\mathop{\mathrm{argmin}}_\beta {\cal H}^{(\mathrm{RS})}(\beta;z,\beta^{(0)},\eta)
\label{eq:replica_one_body_x}
\end{align}
where
\begin{align}
{\cal H}^{(\mathrm{RS})}(\beta;z,\beta^{(0)},\eta)&=\frac{\widehat{\Theta}}{2}\beta^2-(\sqrt{\widehat{\chi}}z+\widehat{\mu}\beta^{(0)}-\eta)\beta+\lambda|\beta|.
\end{align}
We further define
${\cal H}^{(\mathrm{RS})*}(z,\beta^{(0)},\eta)
:={\cal H}^{(\mathrm{RS})}(\beta_{\mathrm{RS}}^*(z,\beta^{(0)},\eta);z,\beta^{(0)},\eta)$.

The minimizer $\beta_{\mathrm{RS}}^*$ takes the soft-thresholding form
\begin{align}
    \beta^*_{\mathrm{RS}}(z,\beta^{(0)},\eta)=\left\{
\begin{array}{ll}
\displaystyle\frac{h_{\mathrm{(RS)}}-\lambda\mathrm{sgn}(h_{(\mathrm{RS})})}{\widehat{\Theta}} & \mathrm{for}~ |h_{(\mathrm{RS})}|>\lambda\\
0 & \mathrm{otherwise}
\end{array}
\right.
\end{align}
where
$h_{\mathrm{RS}}(z,\beta^{(0)},\eta)
=\sqrt{\widehat{\chi}}\,z+\widehat{\mu}\beta^{(0)}-\eta$.
The corresponding minimized value of the function ${\cal H}^{(\mathrm{RS})}$ is given by
\begin{align}
    {\cal H}^{\mathrm{(RS)}*}(z,\beta^{(0)},\eta)=\left\{
\begin{array}{ll}
\displaystyle\frac{\left(h_{\mathrm{(RS)}}-\lambda\mathrm{sgn}(h_{(\mathrm{RS})})\right)^2}{2\widehat{\Theta}} & \mathrm{for}~ |h_{(\mathrm{RS})}|>\lambda\\
0 & \mathrm{otherwise}
\end{array}
\right..
\end{align}

At the extremum, the macroscopic parameters satisfy
\begin{align}
Q&=\int d\eta P_\eta(\eta)\int d\beta^{(0)}\phi^{(0)}(\beta^{(0)})\int Dz(\beta_{\mathrm{RS}}^*(z,\beta^{(0)}))^2\label{eq:Q_RS}\\
\chi&=\frac{1}{\sqrt{\widehat{\chi}}}\int d\eta P_\eta(\eta)\int d\beta^{(0)}\phi^{(0)}(\beta^{(0)})\int Dz\frac{\partial \beta_{\mathrm{RS}}^*(z,\beta^{(0)})}{\partial z}\label{eq:chi_RS}\\
m&=\int d\eta P_\eta(\eta)\int d\beta^{(0)}\phi^{(0)}(\beta^{(0)})\int Dz\beta_{\mathrm{RS}}^*(z,\beta^{(0)})\beta^{(0)}.\label{eq:m_RS}
\end{align}

\subsection{Connection to the state evolution}

For high-dimensional models with a random design matrix, the fixed-point equations of state evolution coincide with those obtained by the replica method. In particular, the macroscopic quantities derived from AMP and state evolution agree with those predicted by the replica analysis, namely
\begin{align}
    Q&=\mathbb{E}_{{\cal D},\bm{\eta}}\left[\frac{1}{p}\sum_{i=1}^p\widehat{\beta}_{\mathrm{AMP},i}^2({\cal D},\bm{\eta})\right]\\
    \chi&=\mathbb{E}_{{\cal D},\bm{\eta}}\left[\frac{1}{p}\sum_{i=1}^ps_i({\cal D},\bm{\eta})\right]=V\\
    m&=\mathbb{E}_{{\cal D},\bm{\eta}}\left[\frac{1}{p}\sum_{i=1}^p\beta_i^{(0)}
    \widehat{\beta}_{\mathrm{AMP},i}({\cal D},\bm{\eta})\right].
\end{align}
Furthermore, the generalization error and training error are expressed as
\begin{align}
    E_{\mathrm{gen}}&=Q-2m+\rho\sigma_\beta^2+\sigma_\xi^2\\
    E_{\mathrm{train}}&=\frac{Q-2m+\rho\sigma_\beta^2+\sigma_\xi^2}{(1+\chi)^2}.
\end{align}
These relations lead to Claim~4.2 in the main text.

In addition, there is the correspondence
\begin{align}
    \Sigma^{-1}=\widehat{\Theta}.
    \label{app_eq:Sigma_correspondence}
\end{align}
Solving \eqref{eq:chi_RS}, we can find that
\begin{align}
    \chi=\frac{\widehat{\rho}}{\widehat{\Theta}}
\end{align}
hence, from \eqref{app_eq:Sigma_correspondence}, we obtain
\begin{align}
    V=\frac{(1+V)\widehat{\rho}}{\alpha},
\end{align}
and transforming this, we obtain eq.(43) in the main text.

These correspondences have been established in a number of previous studies, where the replica method characterizes the typical-case behavior of the estimator. The agreement between state evolution and the replica method supports the validity of the present analysis and provides further justification for the use of AMP in characterizing the typical performance of the estimator.

\bibliographystyle{plainnat}
\bibliography{reference_PrivateLASSO} 

\end{document}